\documentclass[conference]{IEEEtran}     

\usepackage[noadjust]{cite}
\usepackage{graphicx}
\usepackage{subfig}
\usepackage{amsmath}
\usepackage{amssymb}
\usepackage{multirow}
\usepackage{multicol}
\usepackage{balance}
\usepackage{csquotes}
\usepackage{hyperref}
\usepackage{tikz}
\usepackage{booktabs}
\usepackage{comment}
\usepackage{stfloats}

\DeclareMathOperator*{\argmax}{arg\,max}
\DeclareMathOperator*{\argmin}{arg\,min}

\newcommand{\DKL}{\mathbb{D}_{\mathrm{KL}}}
\newcommand{\para}[1]{\vspace{0.5em}\noindent\textbf{#1} }
\newcommand{\mirror}{\textsc{Mirror}}

\newcommand{\mirrorkl}{\textsc{Mirror-Kl}}
\newcommand{\mirrorp}{\textsc{Mirror-P}}
\newcommand{\bc}{\textsc{Bc}}
\newcommand{\nc}{\textsc{Nc}}
\newcommand{\im}{\textsc{Im}}
\newcommand{\sqil}{\textsc{Sqil}}
\newcommand{\human}{\textrm{H}}
\newcommand{\robot}{\textrm{R}}

\title{\LARGE \bf
MIRROR: Differentiable Deep Social Projection for Assistive Human-Robot Communication
}

\author{Kaiqi Chen, Jeffrey Fong, and Harold Soh\\Dept. of Computer Science, National University of Singapore.\\{\small\texttt{\{kaiqi, jfong, harold\}@comp.nus.edu.sg}}}

\begin{document}

\maketitle
\thispagestyle{empty}
\pagestyle{empty}

\begin{abstract}
Communication is a hallmark of intelligence. In this work, we present MIRROR, an approach to (i) quickly learn human models from human demonstrations, and (ii) use the models for subsequent communication planning in assistive shared-control settings. MIRROR is inspired by social projection theory, which hypothesizes that humans use self-models to understand others. Likewise, MIRROR leverages self-models learned using reinforcement learning to bootstrap human modeling. Experiments with simulated humans show that this approach leads to rapid learning and more robust models compared to existing behavioral cloning and state-of-the-art imitation learning methods. We also present a human-subject study using the CARLA simulator which shows that (i) MIRROR is able to scale to complex domains with high-dimensional observations and complicated world physics and (ii) provides effective assistive communication that enabled participants to drive more safely in adverse weather conditions. 
\end{abstract}
\section{Introduction}
\label{sec:intro}

\begin{figure}
    \centering
    \includegraphics[width=0.90\columnwidth]{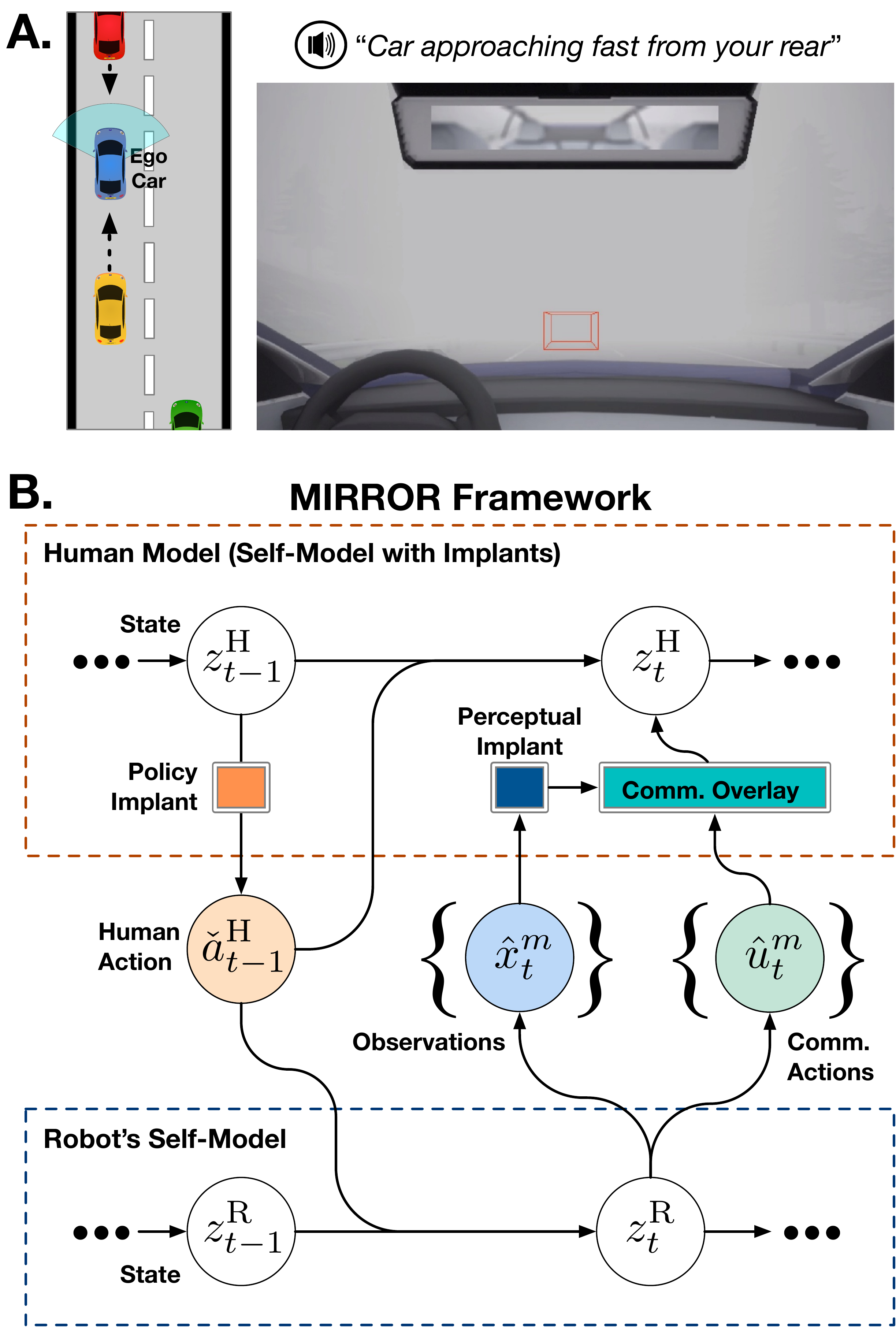}
    \caption{\small Human-Robot Communication Example. (\textbf{A}) A Robot  Assistant needs to provide information to help a human driving the blue car in dense fog. The human has limited visibility and the assistant can highlight cars on a heads-up-display or provide verbal cues (as the human is unable to see highlighted cars that are in the rear). In this scenario, a collision is imminent---the red car in front is slowing down and a yellow car speeding up from the rear. How can the robot determine \emph{what}, \emph{when}, and \emph{how} to inform the user about? Our proposed \mirror{} assistant immediately highlights the red car in front and verbally tells the user about yellow car at the rear. It chooses \emph{not} to tell the human about the green car, whose location has little impact on the human's decision. (\textbf{B}) Our \mirror{} framework is inspired by \emph{social projection}; the robot reasons using a human model that is constructed from its own internal self-model. Strategically-placed learnable ``implants''  capture how the human is different from the robot. \mirror{} plans communicative actions by forward simulating possible futures by coupling its own internal model (to simulate the environment) and the human model (to simulate their actions).}
    \label{fig:main_figure}
\end{figure}

Communication is an essential skill for intelligent agents; it facilitates cooperation and coordination, and enables teamwork and joint problem-solving. However, effective communication is challenging for robots; given a multitude of information that can be relayed in different ways, how should the robot decide \emph{what}, \emph{when}, and \emph{how} to communicate? 

In this paper, we focus on assistive shared-control or teleoperation scenarios. As an example, consider the scenario in Fig. \ref{fig:main_figure} where a robot assistant is tasked to provide helpful information to a human driving the blue car in fog (or to explain the robot's own driving behavior). There are other cars in the scene, which may not be visible to the human. The robot, however, has access to sensor readings
that reveal the environment and surrounding cars. To prevent potential collisions, the robot needs to communicate relevant information --- either  using a heads-up display or verbally --- that intuitively, should take into account what the human driver currently believes, what they can perceive, and the actions they may take. 

Prior work on planning communication methods in HRI typically rely on human models, which are typically handcrafted using prior knowledge (e.g.,~\cite{tabrez2019explanation,buehler2020}) or learned from collected human demonstrations (e.g., ~\cite{reddy2020assisted}). Unfortunately, handcrafted models do not easily scale to complex real-world environments with high-dimensional observations, and data-driven models typically require a large number of demonstrations to generalize well. In this work, we seek to combine prior knowledge with data in a manner that reduces \emph{both} manual specification and sample complexity.

Our key insight is that learning \emph{differences} from a suitable reference model is more data-efficient than learning an entire human model from scratch. We take inspiration from \emph{social projection theory}~\cite{allport1924}, which suggests that humans have a tendency to expect others to be similar to ourselves, i.e., a person understands other individuals using one's self as a reference. This inductive bias can be effective when the agents are similar and may promote cooperation~\cite{krueger2012social,krueger2013social}. From a cognitive perspective, social projection is a heuristic by which we evaluate and predict another agent's behavior.  Likewise, a robot can use its self-model to reason about a human; in our setup, the robot first learns how to perform the task on its own then uses this model to reason about other agents. 

A natural concern is that the human is unlikely to be similar to the robot; indeed, recent work has shown that (model-free) RL agents that are trained to respond to self-policies do not work well with actual humans~\cite{carroll2019utility}. The ways in which the human and robot perceive the world and make decisions are likely to differ. Here, we aim to isolate and learn these differences in a sample-efficient manner. 

We argue that latent state-space models obtained via deep reinforcement learning are well-suited for this purpose --- modern deep variants~\cite{lee2020stochastic} can handle multiple high-dimensional sensory modalities, capture complex dynamics, and are useful for robot decision-making, yet are sufficiently modular to permit structural interpretability~\cite{hafner2019dream,chen2021app}. We exploit this modularity and strategically place \textit{learnable implants} that capture differences in perception and/or policy (Fig. \ref{fig:main_figure}.B.). These implants can be small and relatively unstructured (e.g., neural networks) or specified using prior knowledge about the human (e.g., known cognitive biases or perceptual limitations) with associated parameters that can be quickly optimized via gradient-based learning. 

We call our framework \textbf{M}odel \textbf{I}mplants for \textbf{R}apid \textbf{R}eflective \textbf{O}ther-agent \textbf{R}easoning/Learning (\mirror);  an allusion to the mirror neurons in human brains that are hypothesized to play a role in social projection~\cite{bekkali2021putative,rizzolatti2004mirror}.
Returning to our example, our \mirror{}-enabled robot highlights the red car in front and verbally informs the user about yellow car at the rear. To avoid distracting the user, it chooses \emph{not} to tell the human about the green car, whose location and velocity has little impact on the human's decision and safety. 

These communicative actions are the result of \emph{planning} over the implant-augmented self-model and the robot's internal model (Fig. \ref{fig:mirror_comm}).
Specifically, perceptual implants in the \mirror{} model were used to capture that the human had limited visibility, and could not see highlighted cars in the rear (but could be notified about them via auditory means). To plan forward, the robot samples possible future trajectories using \emph{both} its self-model and the human model, which are coupled together via generated observations and predicted actions (Fig. \ref{fig:main_figure}.B). Unlike recent work~(e.g., \cite{reddy2020assisted,tabrez2019explanation,das2021explainable,lee2019feifeili1}), planning with structured deep \emph{multi-modal} self-models allows the robot to take into account longer-term behavior and choose among rich communication modalities, without having to hard-code environmental or human properties (e.g., belief dynamics). 

Experiments in three simulated domain show that \mirror{} is able to learn human models faster and more accurately compared to behavioral cloning (BC) and a state-of-the-art imitation learning method~\cite{reddy2019sqil}. In addition, we report on a human-subject study ($n=21$) using the CARLA simulator~\cite{dosovitskiy17}, which reveals that \mirror{} provides useful assistive information, enabling participants to complete a driving task with fewer collisions in adverse visibility conditions.

In summary, this paper makes the following contributions:
\begin{itemize}
	\item \mirror{}, a sample-efficient framework for learning human models using deep self-models for initial structure;
	\item A planning-based communication approach that leverages learned world dynamics and human models;
	\item Findings from a human-subject study in the assistive driving domain, showing that \mirror{} provides useful communication that improves task performance. 
\end{itemize}
We believe \mirror{} is a step towards better  data-efficient human models for human-robot interaction; 
to our knowledge, \mirror{} is the first work to demonstrate how deep representation learning during RL can be combined with demonstrations for human modeling and planning. \mirror{} can be used for human-robot communication during robot tele-operation and shared-control settings, and it opens up an alternative path to human-robot collaboration with deep models. To motivate future work, we discuss current limitations and potential research avenues in the concluding section of this paper.  
\section{Background \& Related Work}
\label{sec:related_work}

\mirror{} builds upon the existing literature on human modeling and human-robot communication. Due to its importance, the field of agent communication is large; here, we briefly summarize closely-related work that learn and use human models for human-robot/AI interaction and communication~\cite{tabrez2020,carroll2019utility,choudhury2019utility,mavridis2015}. 

\para{Human Models for HRI.} 
In this work, we focus on model-based methods that explicitly model human behavior. Compared to model-free approaches to HRI, model-based methods tend to make reasonable predictions with far less data~\cite{choudhury2019utility}. Model-based methods can be ``black-box'' in that they make few assumptions about the human and focus on learning a policy function. For example, recent work learns a human policy via imitation learning, followed by a residual policy for shared control~\cite{schaff2020residual}.
In contrast, Theory of Mind (ToM) models incorporate (possibly strong) assumptions about how humans perceive the world and make decisions. For example, a ToM model may assume people are rational and learn in a Bayesian manner~\cite{baker2017rational,ho2021cognitive,soh2019multi}, which is generally not true~\cite{ho2021cognitive}.  

\mirror{} can be seen as a hybrid approach that scaffolds human model learning using the robot's own internal model (obtained using RL). Compared to standard black-box human models, \mirror{} provides additional structure that can ease data requirements. Compared to handcrafted ToM approaches~\cite{buehler2020}, \mirror{} is able to handle high-dimensional observations. \mirror{} is related to recent approaches that focus on capturing human traits, e.g., biases under risk and uncertainty~\cite{kwon2020humans} or action errors due to misunderstood environmental dynamics~\cite{reddy2018you}. However, these approaches typically build on top of hand-crafted ToM models.  

\para{Assistance via Human-Robot Communication.} Enabling robots to communicate with humans has had a long history; early robots in the 1990s (e.g., Polly \cite{horswill1993polly} and RHINO \cite{buhmann1995rhino}) were simple stimulus-response systems. In contrast, modern day robots leverage learning and planning to generate a variety of communication patterns, e.g., legible motion~\cite{dragan2013legibility, dragan2015effects, busch2017learning}, and natural language~\cite{kollar2010toward, thomason2020vision}. 

Recent work has shown that human-robot communication can  improve human task performance~\cite{tabrez2019explanation, unhelkar2020decision}, explain robot errors~\cite{das2021explainable} and calibrate  human-robot trust~\cite{lee2020getting}. However, these approaches typically use hand-specified human models and known environment models. In contrast, \mirror{}
 plans communication actions using \emph{learned} models. \mirror{} is related to prior work on personalized assistive navigation~\cite{ohnbar2018personalized}, but the mechanism differs: \mirror{} adapts implant parameters whilst \cite{ohnbar2018personalized} uses a mixture of expert models. 

\mirror{} is closely-related to Assistive State Estimation (ASE)~\cite{reddy2020assisted} in that both approaches augment user observations to communicate state information. However, there are crucial differences: ASE assumes known dynamics and perceptual models to compute the near-optimal human policies, while MIRROR uses learned dynamics and implant models. In addition to differences in the human model, ASE generates observations that minimize the KL divergence between the (predicted) user's beliefs and the assistant's beliefs. Instead, \mirror{} forward simulates possible futures using its internal models, and plans communication to maximize task rewards while minimizing communication costs.

\section{Model Implants for Rapid Reflective Other-agent Reasoning/Learning (\mirror{})}
\label{sec:mirror}

Our problem setting is one of assistance: a (human) user is acting in a partially-observable environment to maximize rewards. The assistive robot's goal is to help the user achieve their objective. The robot may receive different observations from the environment and can modify the user's observations to provide additional information. We seek to derive an effective assistant. At a high level, we will imbue the robot with a structured model of the human that can be adapted with data. After learning, the robot plans using its own internal (self) model and the human model to communicate valuable information. We first detail the robot's underlying self-model, then describe the human model (specifically the implants), and finally, how communication can be achieved using both the self and human models.

\subsection{Self Model: Multi-Modal Latent State-Space Model}
\label{sec:mssm}

\begin{figure}
    \centering
    \includegraphics[width=0.80\columnwidth]{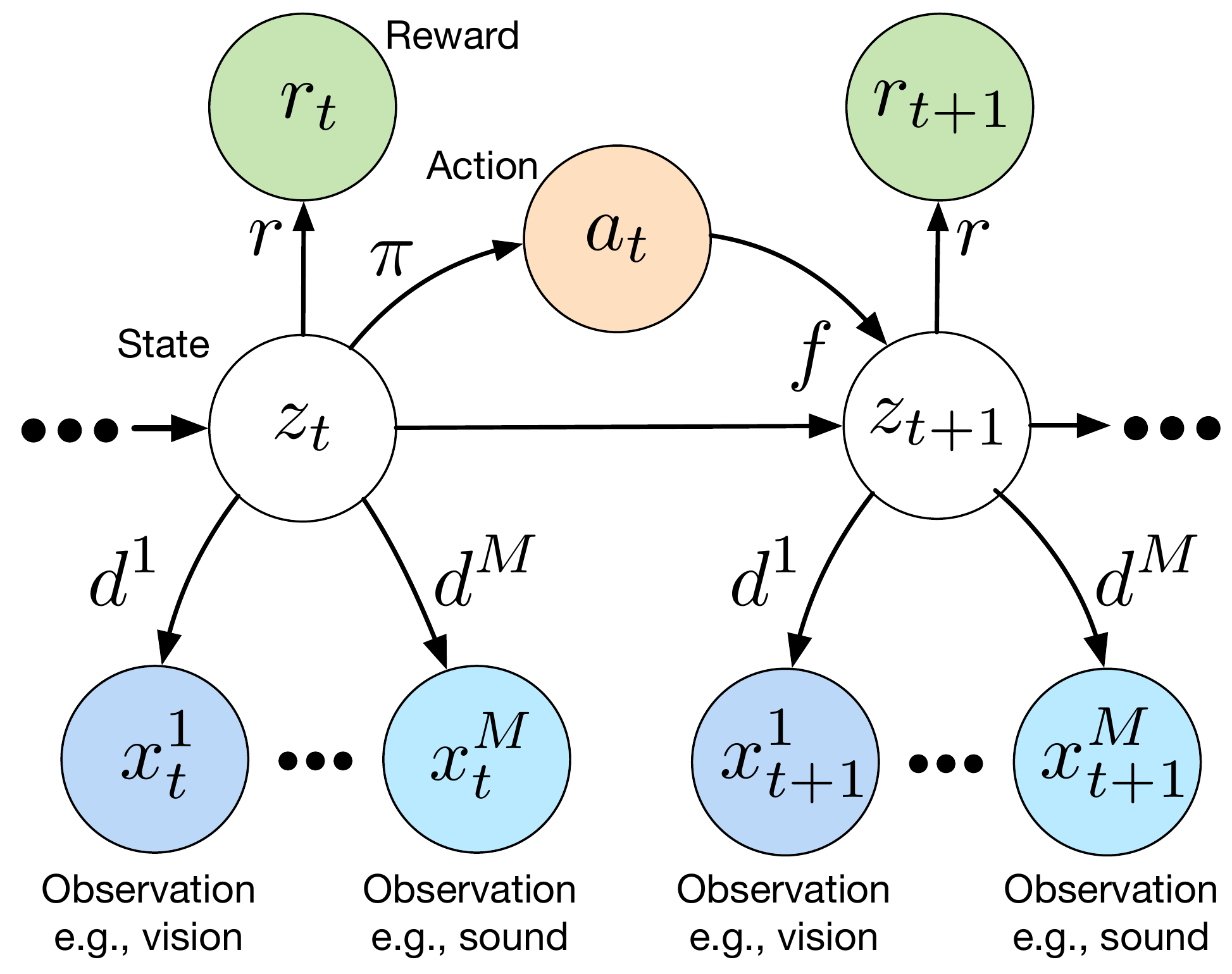}
    \caption{\small \mirror{}'s self-model is a multi-modal latent state-space model (MSSM). In the above, circle nodes represent random variables and shaded nodes are observed during learning.}
    \label{fig:pgm}
    \vspace{-1em}
\end{figure}

\para{Model Structure.} In \mirror, the robot's self-model is a multi-modal state-space model (MSSM)~\cite{chen2021app} (Fig. \ref{fig:pgm}). Intuitively, the MSSM models an agent that is sequentially taking actions in a world and receiving rewards and multi-modal observations. The observation $x^m_t$ at time $t$ for $m=1, \dots, M$ sensory modalities are generated from the latent state $z_t$. The model assumes Markovian transitions where the next state is conditioned upon the current state and the action $a_t$ taken by the agent. Upon taking an action, the agent receives reward $r_t$. In general, the reward can also be conditioned upon the action and next state. Given the graphical structure in Fig. \ref{fig:pgm}, the MSSM's joint distribution factorizes as:
\begin{align}
    &p_\theta(x^{1:M}_{1:T},r_{1:T},z_{0:T},a_{1:T-1}) = \prod_{t=1}^{T} \left[\prod_{m=1}^{M}p_\theta(x^m_{t}|z_{t})\right]\nonumber \\
    &p_\theta(r_{t}|z_{t})p_\theta(z_{t}|z_{t-1},a_{t-1})p_\psi(a_{t-1}|z_{t-1})p(z_0)
    \label{eqn:basicfactorization}
\end{align}
where $\theta$ are the world model parameters and $\psi$ are the policy parameters, $x^{1:M}_{1:T}$ denotes all observations from $t=1,\dots,T$, and similarly for $z_{0:T}$, $a_{1:T-1}$, and  $r_{1:T}$. In this work, each of the factorized distributions are modelled using nonlinear functions: 
\begin{itemize}
   \item Transitions: $p_\theta(z_{t}|z_{t-1},a_{t-1}) = p(z_{t}| f_\theta(z_{t-1},a_{t-1}))$
   \item Observations:  $p_\theta(x^m_{t}|z_{t}) = p(x^m_{t}| d^m_\theta(z_{t})) $ 
   \item Rewards: $p_\theta(r_{t}|z_{t}) = p(r_{t}| r_\theta(z_{t})) $
   \item Policy: $p_\psi(a_{t}|z_{t}) = p(a_{t}| \pi_\psi(z_{t}))$
\end{itemize}
where $f_\theta$, $d^m_\theta$, $r_\theta$, and $\pi_\psi$ are neural networks. Note the slight abuse of notation; we denote both the reward function and the reward random variable as $r$, where the meaning should be clear from context. Depending on the application, the policy $\pi_\psi$ may be deterministic or stochastic. We write $\pi_\psi(a_{t}|z_{t}) = p_\psi(a_{t}|z_{t})$ to refer to the latter case. 

\para{Dynamics Learning.} Given trajectories of the form $\tau = \left\{(x^{1:M}_{t}, a_{t}, r_{t})\right\}_{t=1}^{T}$, we seek to learn the parameters $\theta$. Because maximum likelihood estimation is intractable in this setting, we optimize the evidence lower bound (ELBO) under the data distribution $p_d$ using a variational distribution $q_\phi$ over the latent state variables $z_t$, 
\begin{align}
\mathbb{E}_{p_d}[\mathcal{L}_e] \leq \mathbb{E}_{p_d}[\log   p_\theta(x^{1:M}_{1:T},r_{1:T}|a_{1:T-1})]
\end{align}
where 
\begin{align}
\label{eqn:ELBO}
     \mathcal{L}_e = & \sum_{t=1}^{T}\Big(\displaystyle  \mathop{\mathbb{E}}_{q_{\phi}(z_{t})}\left[\sum_{m=1}^M\log p_{\theta}(x^m_{t}|z_{t})\right] +\displaystyle\mathop{\mathbb{E}}_{q_{\phi}(z_{t})}\left[\log p_{\theta}(r_{t}|z_{t})\right]  \nonumber\\
    &- \displaystyle\mathop{\mathbb{E}}_{q_{\phi}(z_{t-1})}\left[\DKL\left[q_{\phi}(z_{t}) \| p_{\theta}(z_{t}|z_{t-1},a_{t-1})\right]\right]\Big)
\end{align}
The first two terms in the ELBO are reconstruction terms and the  Kullback-Leibler (KL) divergence term encourages consistency between the variational distribution and the transition dynamics. In this work, $q_\phi$ is an inference network,
\begin{align}
    q_\phi(&z_t | z_{t-1}, a_{t-1}, x^{1:M}_{t}) = \nonumber \\ &\mathcal{N}(g^\mu_\phi(z_{t-1}, a_{t-1}, x^{1:M}_{t}), g^\Sigma_\phi(z_{t-1}, a_{t-1}, x^{1:M}_{t}))
\end{align}
where $g^\mu_\phi$ and $g^\Sigma_\phi$ are neural networks that give the mean and the covariance of the Gaussian latent state variable, respectively. When observations are missing, we apply a simple masking operation similar to prior work~\cite{lipton2016directly}, but alternative approaches such as a product-of-experts~\cite{hinton2002training} can be used as well (with corresponding changes to the inference network structure). The inference network $q_\phi$ will continue to play an important role in \mirror{} to facilitate fast inference and planning during communication. 

\para{Policy Learning.} 
Given the latent states $z_{t}$ sampled from the inference network, we leverage RL to learn optimal policies. Our approach is based on Stochastic Latent Actor Critic~\cite{lee2020stochastic}. In our human-subject study involving the assistive driving task, we train a critic network $Q_{\psi}(z_{t}, a_{t})$ and an actor network $\pi_{\psi}(a_t|z_t)$ using Soft-Actor Critic (SAC)~\cite{haarnoja2018soft}. In our simulated experiments involving gridworld environments, we train a Q-network $Q_{\psi}(z_t, a_t)$ using Deep Q-Learning (DQN)~\cite{mnih2013dqn}.

\para{Practical Aspects.} We find that eventual human model performance is significantly improved when the self-model is trained with data-augmentation~\cite{laskin2020reinforcement} and random dropouts across the sensory modalities~\cite{wu2018mvae,zhi2020factorized}. For example, we randomly drop LIDAR range readings in each training batch when training our robot in CARLA. We hypothesize that to generalize appropriately, the model should be trained with data that approximates the different ways observations can appear to a human. As an added bonus, this training method often resulted in more robust internal world models and policies.

\subsection{Learning Human Models via \mirror{}}
\label{subsec:mirror}
\begin{figure}
    \centering
    \includegraphics[width=0.9\linewidth]{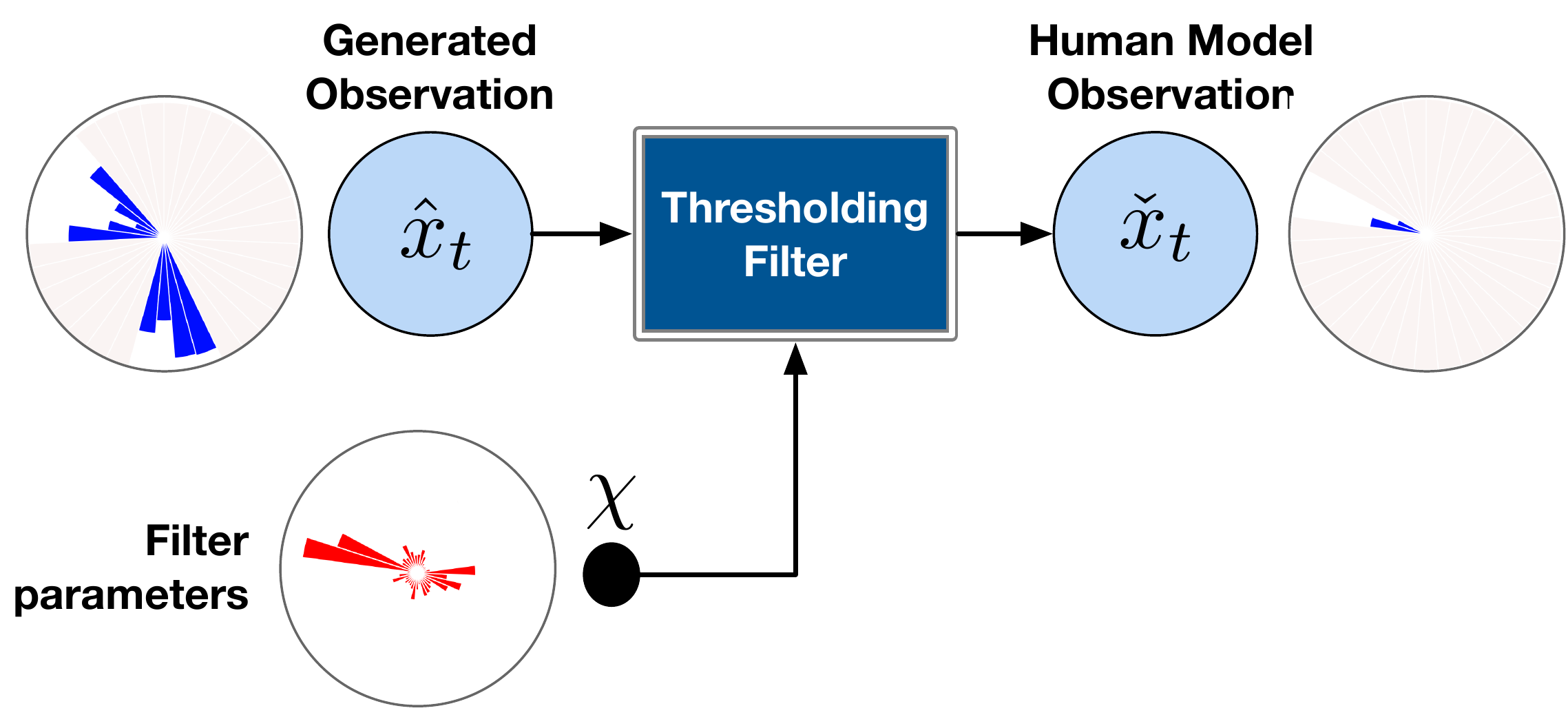}
    \caption{{\small An Example Thresholding Filter as a Perceptual Implant. The (generated) range observations (blue bars) are passed through a thresholding filter to eliminate observations that are beyond a parameterized distance (red bars) in each segment from the human.}}
    \label{fig:perceptual_filter}
\end{figure}

In \mirror{}, the human model is identical to the robot's self-model \emph{except} for implants that are injected to change the model's behavior.
To distinguish variables in the human and robot models, we will use the superscript H and R to refer to the human or robot, respectively. For example, the human action at time $t$ is $a^\human_t$. Furthermore, we will denote generated/predicted variables by a hat accent, e.g., $\hat{a}^\human_t$, and variables that are changed by an implant with a check accent, e.g., $\check{a}^\human_t$. 

\para{Model Implants.} Once the self-model is trained, we augment it with implanted functions $h(\cdot)$. In this work, we examine two implant classes: 
\begin{itemize}
    \item \textbf{Perceptual implants} model how humans perceive the world by changing the observation $x_t$. In several of our experiments, we use a threshold mask/filter implant that models that the human can only see objects within a certain range (Fig. \ref{fig:perceptual_filter})
    \item \textbf{Policy implants} model how the human acts differently from the robot by changing the policy. Simple implants can simply add noise to increase policy entropy. In this work, we add a state-dependent residual $\delta(z_{t}^\human)$ term to the parameters of the policy distribution $\pi$. As an example, for the Gaussian policies  $\mathcal{N}\left(\mu(z_t^\human), \sigma^2(z_t^\human)\right)$ used in our CARLA experiments, we add $\delta(z_{t}^\human)$ to the mean of the Gaussian such that the augmented policy distribution follows $\mathcal{N}\left(\mu(z_t^\human) + \delta(z_t^\human), \sigma^2(z_t^\human)\right)$. For our discrete policy $\mathrm{Cat}(K, \mathbf{p})$, we add $\delta(z_{t}^\human)$ to the distribution parameters $\hat{\mathbf{p}}=\sigma(\mathbf{p}+\delta(z_t^\human))$, where $\sigma$ is softmax function. The residual is parameterized by a small neural network. 
\end{itemize}
 
The above provides a flavor of what is possible and is not exhaustive. For example, to model humans who are slow to change their beliefs, we could implant a low-pass filter $z^\human_{t+1} = \alpha z_t + (1-\alpha) f(z_t, x_t, a_t)$ where $\alpha$ is a learnable parameter. We leave exploration of other implants to future work.

\para{Learning Implant Parameters.} Given an implant $h_\chi$ parameterized by $\chi$, we can learn $\chi$ by minimizing the following loss given data:
\begin{align}\label{eqn:implant_learning}
   \argmin_\chi \ &\mathcal{L}(\chi) = - \sum_{t=1}^T \mathbb{E}_{z_{t}\sim p(z_{t}| x^{1:M}_{1:t}, a_{1:t-1}, \chi)} \left[\log \pi(a_{t}|z_{t})\right] \nonumber\\ 
    & \quad - \lambda \log p(\chi)
\end{align}

where $\lambda$ is a regularization hyperparameter that controls the strength of an optional prior.
Intuitively, this loss optimizes the likelihood of observing a human's actions given the self-model and the implant parameters. 
In our work, we approximate $p(z_{t}| x^{1:M}_{1:t}, a_{1:t-1})$ using the learnt inference network
and perform stochastic gradient descent by sampling $z_{t}$. Note that \emph{only} the implant parameters are modified when learning from human data.

\subsection{Human-Robot Communication with \mirror{}}

\begin{figure}
    \centering
    \includegraphics[width=0.85\columnwidth]{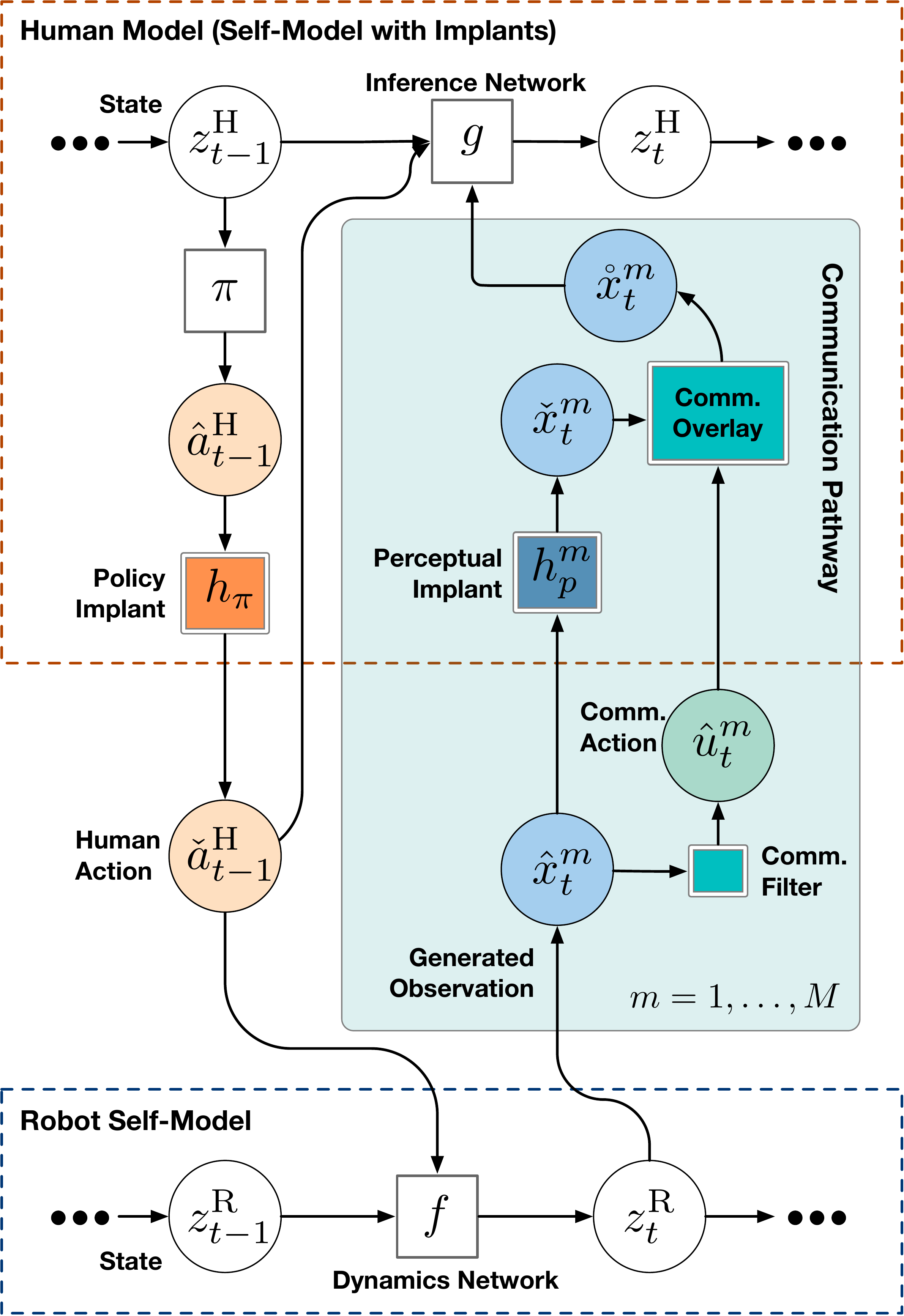}
    \caption{\small Human-Robot Communication via Forward Simulation. In brief, the robot's self-model is used to simulate the environment and the human model is used to simulate actions. Be leveraging the learnt dynamics network $f$ and inference network $g$, the two models are coupled and used to ``imagine'' possible futures. The communication pathways (in teal) serves to link the robot's communication actions $\hat{u}^m_t$ (filtered generated observations) to the human model's observations. The multi-modal models support $M$ potential communication pathways and the robot may choose one or more of these pathways at any given time step. 
    Please see text for additional details.}
    \label{fig:mirror_comm}
\end{figure}

Next, we turn our attention to how the implanted self-model can be used for human-robot communication. The key idea is to plan using both learned models; we couple the robot's self-model and the human model together via generated robot observations and communication actions, and predicted human actions (Fig. \ref{fig:mirror_comm}). 

\para{Communication Pathways.} To enable the robot to influence the human, we must first establish a means of communication between the two agents. Similar to \cite{reddy2020assisted}, we assume the robot can change the human's observations of the world, e.g., by providing some visual cues or alerting the user verbally. A communication pathway for modality $m$ is illustrated in Fig. \ref{fig:mirror_comm} (teal block) where the human's observation $\check{x}^{m}_t$ and the robot's communication action $\hat{u}^{m}_t$ are combined via an overlay function to yield a ``combined'' observation $\mathring{x}^m_t = v(\check{x}^{m}_t, \hat{u}^{m}_t)$. 
As an example, for images, we can replace selected pixels of $\check{x}^{m}_t$ with corresponding pixels from $\hat{u}^{m}_t$.

Communication actions $\hat{u}^{m}_t$ are actually observations that are generated from the robot belief state $\hat{x}^{m}_t \sim p( \hat{x}^{m}_t | d^m(z^\robot_t))$ and passed though a communication filter; for example, we can use simple masks filters (parameterized by $\omega_{t}$) to filter away irrelevant information about the robot's belief. Note that a communication action can be generated for each modality $m$, which offers the robot multiple ways to communicate with the human.

\para{Forward Imagination.} Given our robot model, human model, and communication pathway, we now seek to sample future trajectories. We begin at time-step $t-1$ where we have samples from the robot's belief over the world state $z^\robot_{t-1}$ and the (predicted) human belief $z^\human_{t-1}$. We take a step from $t-1$ to $t$ to obtain $z^\robot_t$ and $z^\human_{t}$. This can be accomplished in seven steps:
\begin{enumerate}
    \item Sample the human's action, $\check{a}^\human_{t-1} \sim  h_\pi(\pi(z^\human_{t-1}))$.
    \item Forward simulate using the world dynamics $z^\robot_{t} \sim p(z^\robot_{t}| f(z^\robot_{t-1}, \check{a}^\human_{t-1}))$;
    \item Generate the observations $\hat{x}^{1:M}_{t} \sim p(\hat{x}^{1:M}_{t}| d^m(z^\robot_{t}))$;
    \item Obtain the human observation using the perceptual implants $\check{x}^{m}_{t} = h^m_p(\hat{x}^{m}_{t})$ across the modalities;
    \item Obtain the communication action using the communication filter $\hat{u}^{m}_{t} = w_\omega(\hat{x}^{m}_{t})$ for each modality $m$;
    \item Combine the communication action and the observation to obtain $\mathring{x}^m_t = v(\check{x}^{m}_t, \hat{u}^{m}_t)$;
    \item Finally, sample the human's belief state using the inference network $z^\human_{t} \sim q(z^\human_{t}| z^\human_{t-1}, \check{a}^\human_{t-1}, \mathring{x}^{1:M}_t)$.
\end{enumerate}

Given the human's action (the observed action if available or generated by the human policy), the forward dynamics function gives the next state  $z^\robot_{t} \sim p(z^\robot_{t}| f(z^\robot_{t-1}, a^\human_{t-1}))$.  Given $z^\robot_{t}$, we generate observations $\hat{x}^{m}_{t}$ for modalities $m = 1, \dots, M$, which are modified by the perceptual implants to yield $\check{x}^{m}_{t}$. The modified observations $\mathring{x}^{1:M}_t$ are used to update the human's beliefs using the inference network: $z^\human_{t} \sim q(z^\human_{t}| z^\human_{t-1}, \check{a}^\human_{t-1}, \mathring{x}^{1:M}_t)$. This process can be iterated to forward propagate the belief states up to a future horizon $T$.

\para{Planning for Communication.} Given the forward simulation and communication pathways above, the robot can optimize communication to maximize task rewards while minimizing costs: 
\begin{align}
\argmax_{\omega_{0:T}} J = \mathrm{E}_{z_{0:T}|\omega_{0:T}}\left[\sum_{t=0}^T {\gamma^t\left(r(z_t)  - C(\omega_{t})\right)} \right]
\end{align}
where $\omega_{0:T}$ is the parameters of our communication filter for time-steps 0 to $T$, $\gamma$ is the discount factor, ${r}$ is the task reward function\footnote{In principle, this task reward function may differ from the reward function that the model was trained with, but we leave such experiments to future work.}, and $C$ is the cost function. The expectation is taken with respect to the trajectories $\tau$ under the models and filter parameters $\omega_{0:T}$. Various methods can be used to optimize $J$. In our work, we use the cross-entropy (CE) method~\cite{rubinstein1999cross}, and re-plan at each time-step. Note that real observations are obtained after each step, which are used to update the beliefs of the robot and human models. 

\begin{figure*}
    \centering
    \includegraphics[width=0.8\textwidth]{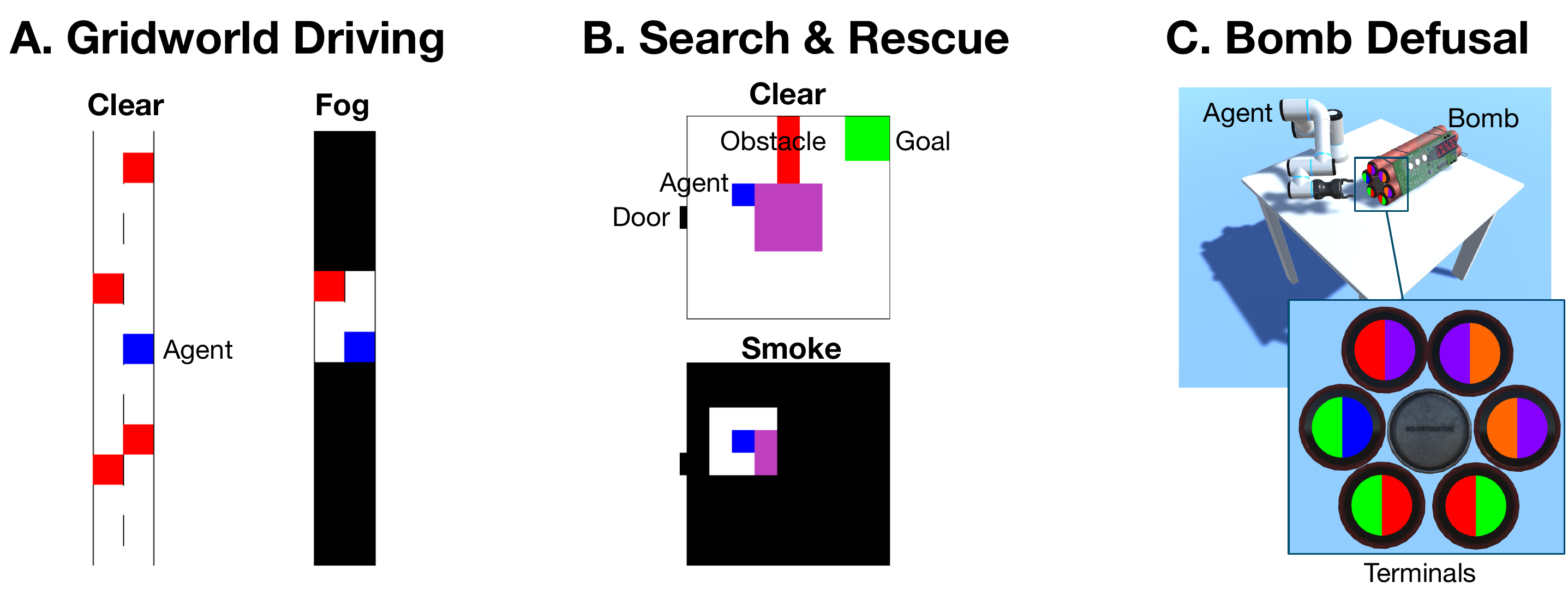}
    \caption{\small Experimental domains. (\textbf{A}) Gridworld Driving where the blue vehicle is moving on a road at constant speed and has to avoid the other red vehicles. In the Fog setting, visibility is reduced; the black region indicates areas not visible to the human. (\textbf{B}) A Search-\&-Rescue task where the agent (blue box) starts at the door and is tasked to rescue a victim at the green goal and bring them back to the door. The obstacle in red can appear in either the top or the bottom path, and the victim's position is randomly initialized in one of three potential positions. In the Smoke variant, visibility is reduced to a small region around the human (\textbf{C}) A Bomb Defusal game where a teleoperated robot has 15 seconds to disarm the bomb by pressing three buttons (one in each stage). The correct button at each stage depends on six visible ``terminals'' (which change after each button press), the bomb type (not visible to the human, but detectable by the robot) and game rules. The rules differ slightly between the robot training environment and the test environment. The human, who has access to the updated rules, has to confirm the robot's selection. This domain features asymmetric sensing and knowledge/policy between the human and robot. In all three domains, the assistive agent/robot has to communicate relevant information to help the human agent complete the task successfully.}
    \label{fig:exp1_domains}
\end{figure*}

\section{Experiments}
\label{sec:exp_human_model}
In the following, we describe experiments with simulated humans (created using real human data) in a simplified driving task, a search-\&-rescue, and a bomb defusal game (Fig. \ref{fig:exp1_domains}). Using simulated humans allowed us to compare the learned models against ground truth models with varying amounts of training data, and enabled us to test communication.
Our primary hypothesis was that leveraging the self-model and learning the implant parameters allows \mirror{} to better model human behavior using fewer samples. In particular, we posit that using \mirror{} results in better communication during transfer (e.g., from clear weather to dense fog/smoke where visibility is reduced). We focus on the main results and relegate details (e.g., domain parameters) to the Appendix. Source code is available at \href{https://github.com/clear-nus/mirror}{https://github.com/clear-nus/mirror}. 

\subsection{Experimental Setup} 

\para{Domains and Communication Modalities.} Our setup comprised a primary human agent who is attempting to complete a task, and an assistive robot who is able to reveal information via communication. Figure \ref{fig:exp1_domains} summarizes the three domains used in our experiments. The Driving task is similar to those used in prior work in HRI (e.g., \cite{choudhury2019utility}) where state information is available to the agent. The Search-\&-Rescue domain is a more complex gridworld environment, where the robot can only observe raw image data and textual descriptions of the position of the victim and obstacles. 
In these two domains, there is a ``transfer'' setting (fog/smoke) where the human's visual perception is degraded, i.e., it doesn't receive any state information beyond its field of view. For the Bomb Defusal game, the transfer setting is different in that the robot's performance is diminished; the rules dictating which button to press is different from the rules it was trained with. As such, the communication explains the robot's choice and the human has to verify. 

In all three domains, there are two communication modalities: visual and verbal. The total cost for each modality is quadratically related to the number of items communicated. Verbal communication is more costly, but can reveal information that is not visually apparent (e.g., vehicles in the rear or the bomb type). For more information, please see Table \ref{tbl:comm_modalities} in the Appendix. 

\para{Simulated Humans.} For each domain, we created simulated human agents using human data; we collected data from 10 real humans playing multiple rounds (24 to 36 rounds depending on the domain). The data from each person was used to train an agent model for that individual. 
For example, in the Driving domain, we used Maximum Entropy IRL~\cite{ziebart2008maximum} to learn a reward function from the collected trajectories. Similar to prior work~\cite{choudhury2019utility}, we use as features (i) the distance to the center of the road, (ii) distances to the other cars, and (iii) the driver's action.  Given the learnt reward function, the simulated human agent perceives state information (up to its perceptual capabilities) and plans at each time-step to maximize its discounted cumulative reward up to a specified horizon. Further details about the other simulated agents are in Sec. \ref{app:real_sim_humans} of the Appendix. 

\para{Compared Methods.} In total, we compare five different human modeling methods: 
\begin{itemize}
    \item Ideal Model (\im): This baseline uses the same model as the simulated human and thus, represent ``ideal'' models that are typically unavailable in practice. These serve as an upper-bound on performance.
    \item Behavioral cloning (\bc): a black-box policy learnt via supervised learning on observed trajectories. 
    \item Soft Q-Imitation Learning (\sqil)~\cite{reddy2019sqil}: A state-of-the-art method that trains a policy via RL to match the human demonstrations. \sqil{} represents a class of imitation learning methods that have access to the environment\footnote{In preliminary experiments, we also tested GAIL~\cite{ho2016generative} but \sqil{} outperformed GAIL on all our tests.}.
	\item \mirror{}: With a perceptual implant and a policy implant. See Appendix (Sec. \ref{sec:exp mirror_implants}) for implant details. 
	\item \mirrorp{}: A \mirror{} variant with the perceptual implant only.   
\end{itemize}
It is important to note that \bc{}, \sqil{}, and \im{} are \emph{aware} of what the human can perceive; we provided the ground truth human observations. This is \emph{not} the case for the \mirror{} models that perceive what the robot  observes and had to learn what the human could observe. The \mirror{} self-models were trained to perform the task in the original domain, and only the implants were adapted using the simulated human data (in both the original and transfer settings). 

To train \bc{} and \sqil{} in the transfer domains, we combined both the original and transfer data, e.g., if 10 trajectories were used to train \mirror{}, we used 20 trajectories (10 original, 10 transfer) for \bc{} and \sqil{}. This was necessary to obtain reasonable performance for these methods. Each method was used to train a neural network policy (comprising GRUs with fully-connected layers) with early-stopping on a validation set. 
We varied the number of layers and report best results for the  baseline models. 

\begin{figure}
    \centering
    \includegraphics[width=1.0\columnwidth]{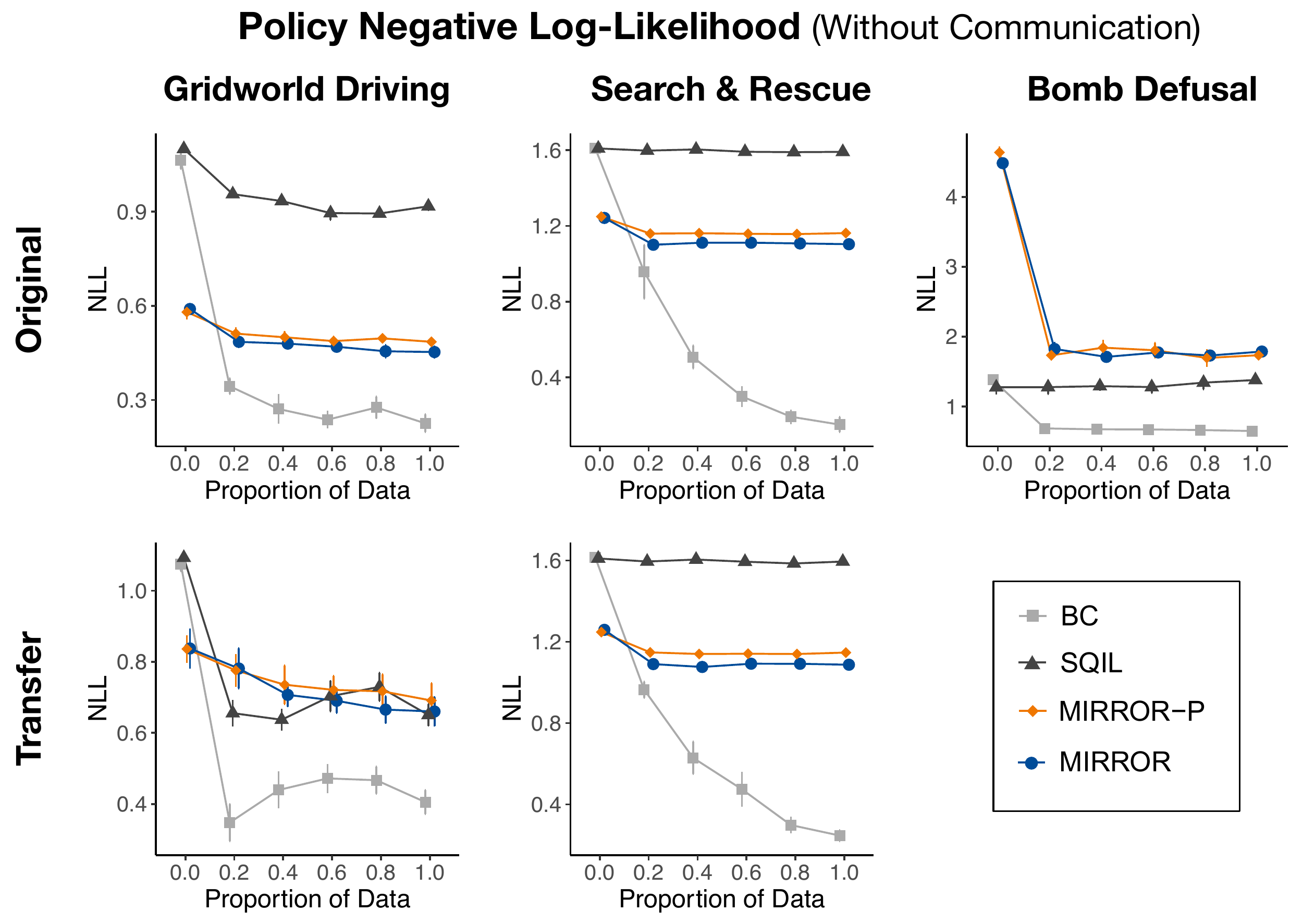}
    \caption{\small Policy Negative Log-Likelihood (NLL) in the different domains after training with different proportions of data. Lower scores indicate better fit to the data. In both the original and transfer settings, \bc{} achieved the best scores. However, this did not translate into good performance during communication. See main text for details.}
    \label{fig:gridworld_nll}
    \vspace{-1.5em}
\end{figure}

\begin{figure}
    \centering
    \includegraphics[width=1.0\columnwidth]{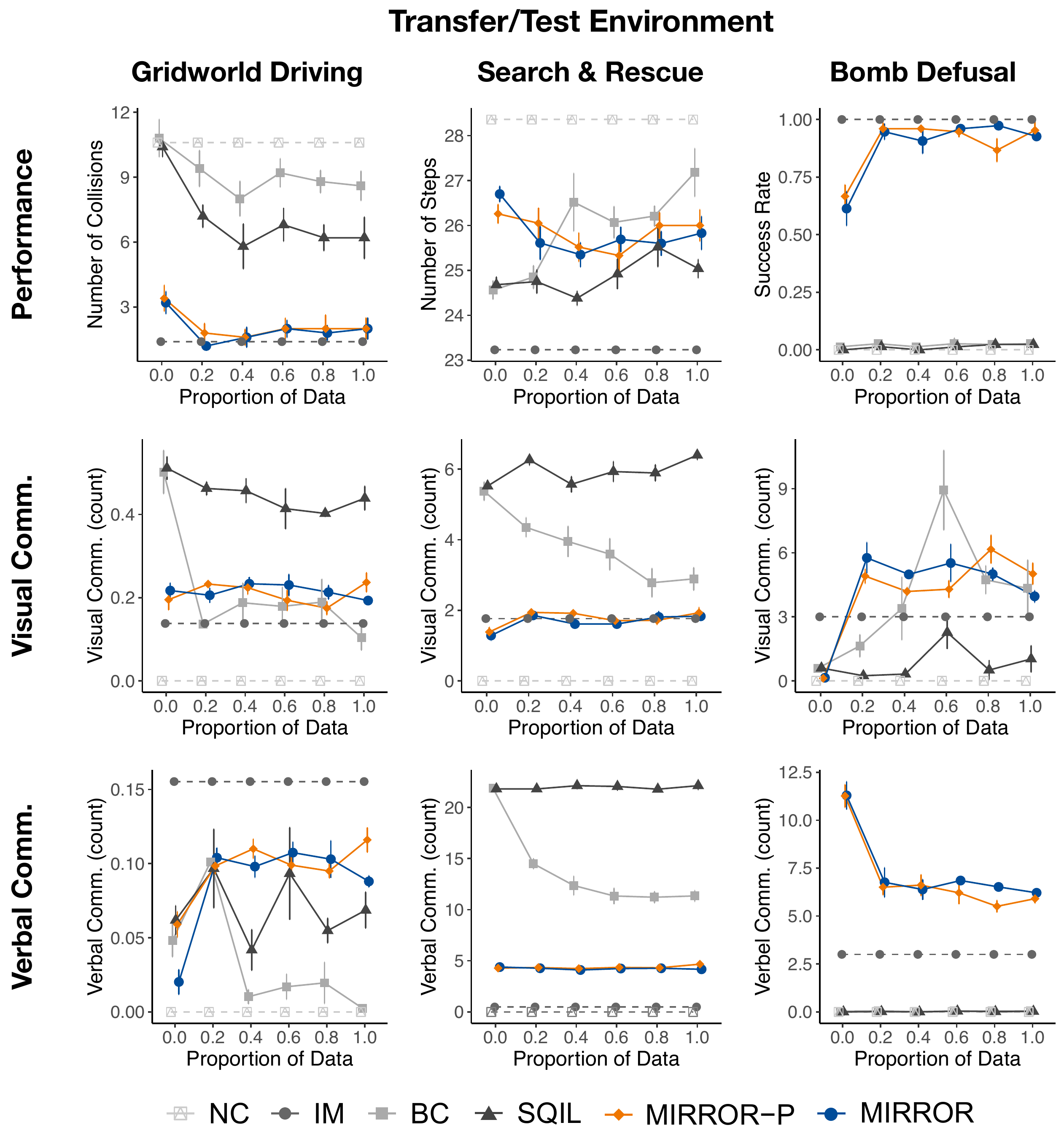}
    \caption{\small Task Performance and Communication Amount in the Transfer/Test Environment. \nc{} indicates agents that received no communication. In all domains, the simulated human's performance improved given communication. For the Search \& Rescue Task, the \sqil{} agents slightly outperform the \mirror{} agents but at far higher communication cost. In the Driving and Bomb Defusal tasks, the best performing simulated agents were those that were paired with the assistive \mirror{} agents. The performance was achieved with reasonable amounts of communication, close to that of the ideal models.}
    \label{fig:gridworld_comm_results}
    \vspace{-1.5em}
\end{figure}

\subsection{Results and Analysis}

Overall, we find that \mirror{} is able to learn better human models with fewer samples compared to \bc{} and \sqil{}. We initially compared the negative log-likelihood (NLL) of the trained policies (on a test set) with different amounts of training data (Fig. \ref{fig:gridworld_nll}). Interestingly, \bc{} achieved the best NLL scores, indicating that the \bc{} models best fit the data and may be good proxy human models for planning. However, these scores are misleading. 

The \mirror{} models were far more effective than \bc{} for communication planning. Fig. \ref{fig:gridworld_comm_results} summarizes the performance of the simulated human agents with communication from the compared methods, along with the amount of communication provided. In all domains, the agents that received communication outperformed the agents that did not (\nc).
For the Gridworld Driving and Bomb Defusal tasks, the simulated humans achieved significantly better performance scores with reasonable amounts of communication from the \mirror{} assistants. \mirror{} learned rapidly---good models could be obtained with 20\% of the data (4 trajectories).  For the Search-\&-Rescue task, \sqil{} obtained the best performance scores (about 1-2 steps better than \mirror), but only by using large amounts of verbal communication. Qualitatively, the \sqil{} agent would reveal almost the entire map, but with no discernible pattern to the communication. Moreover, providing more training data did not reduce the verbal communication provided. On the other hand, the \mirror{} models revealed information more selectively, often focusing on the locations of the goal item and the obstacle.

\section{Human-Subject Experiments}

\begin{figure}
    \centering
    \includegraphics[width=0.9\columnwidth]{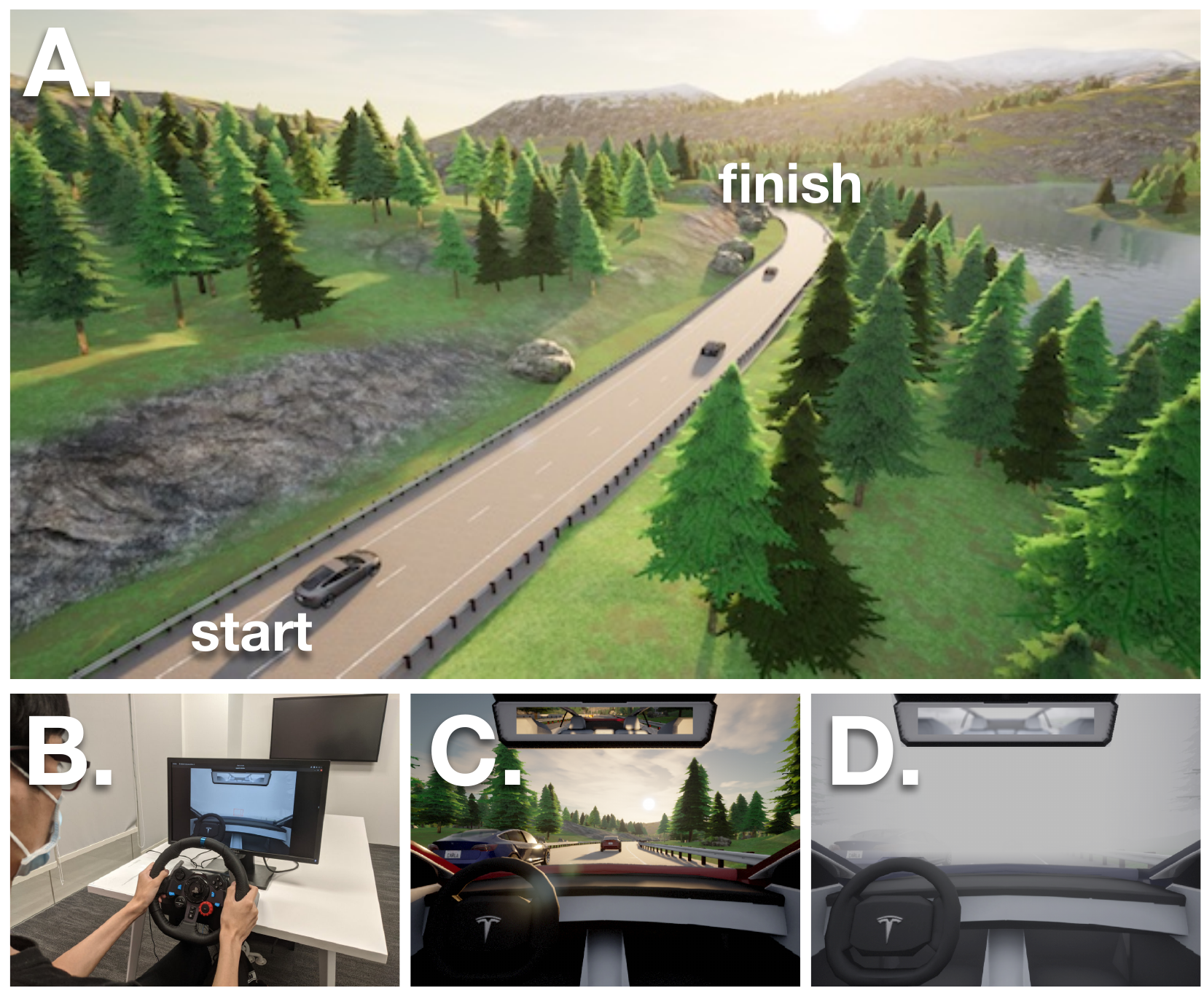}
    \caption{\small CARLA Experiment Setup. (\textbf{A}) The stretch of highway that participants drove along. (\textbf{B}) Participants drove the simulated car using a steering wheel with accelerator and brake pedals. (\textbf{C}) and (\textbf{D}) show the difference in visibility in clear and foggy weather. Both cars are  visible in the clear setting. In the fog setting, the car on the left is visible, but the car in the front can barely be seen.}
    \label{fig:carla}
\end{figure}

In this section, we report on a second set of experiments designed to test whether \mirror{} is able to provide useful information to human users in a more realistic setting. 
We use CARLA \cite{dosovitskiy17}, a modern driving simulator (Fig \ref{fig:carla}). Unlike the previous gridworld experiments, the CARLA environment has continuous state and action spaces, with realistic dynamics and visuals. Our main hypotheses were that (\textbf{H1}) planning with implanted self-models would yield more helpful communication than planning with behavioral cloning models and (\textbf{H2}) planning to optimize task rewards and communication costs would lead to less redundant communication compared to belief matching. Our study was approved by our institution's ethical review board. 

\subsection{Experimental Setup}

\para{Task Description.} Participants interacted with our assistive driving agents in a highway driving task under dense fog (see Fig. \ref{fig:carla}.C and \ref{fig:carla}.D for a comparison between clear and adverse weather conditions). The goal was to drive along a stretch of CARLA's Town04's two-lane carriageway from the starting position to the destination, while navigating through a normal highway traffic. The other vehicles may slow down or speed up, and due to poor visibility, would not be able to avoid the participant's vehicle. The participant had to actively avoid other vehicles along the way. 

\para{Assistive Communication.} The car is equipped with a semantic LIDAR and a driving assistant that can provide both visual and verbal cues. Specifically, the agent could highlight selected vehicles through visual bounding boxes and/or provide informative speech (as previously shown in Fig. \ref{fig:main_figure}.A). The visual bounding boxes can be generated when the other cars enter the range of LIDAR detection range. Verbal communication comprised of free text generated by OpenAI's GPT-2 language model~\cite{radford2019gpt} (fine-tuned for our task). The system  is capable of informing participants when a car approaches or slows down, as well as when no cars are detected in a specific direction. The robot is able to observe 36 LIDAR beams along three angles (108 beams in total), the velocity of the ego car, distances to the center of the lane, and the relative curvature of the road 10 meters ahead.

\para{Compared Methods.} Based on our previous experiments, we trained \mirror{} with both perceptual and policy implants. We compared four conditions:
\begin{itemize}
	\item No Communication (\nc): the human receives no assistance.
	\item Behavioral Cloning (\bc): the human model is trained using behavioral cloning. The model is provided with both expert demonstrations in clear weather, and participant data in the foggy weather.
	\item \mirror{}: our \mirror{} method that first trains a self-model via RL, then learns the implants from demonstrations, and plans communication.
	\item \mirrorkl{}: a \mirror{} variant that does not plan but  minimizes the KL-Divergence between the human's mental state and the robot's belief, similar to \cite{reddy2020assisted}. Unlike \cite{reddy2020assisted}, \mirrorkl{} uses a learnt implant model and is capable of multi-modal communication.
\end{itemize}
We were unable to use \sqil{} due to experimental time constraints; training on the demonstrations with \sqil{} required several hours.  

\para{Participants.} A total of 21 participants (mean age = 23.2, 10 females) were recruited from the university community. The experiment was designed to be within-subjects with all 21 participants in each condition.

\para{Procedure.} Participants entered the lab and were briefed about the task. They then engaged in two practice trials; the first trial involved driving freely along the highway in clear weather conditions, and the second trial involved three rounds of the driving task under dense fog conditions without any assistive communication.
Thereafter, they performed a total of 24 rounds, with the first 6 rounds without any assistive communication, followed by 18 rounds with three different agents (six rounds per agent). The data from the first 6 rounds were used to train the models and the order of the three agents was counterbalanced. Participants could choose to take a one minute rest after every 6-th round to reduce fatigue. 

\para{Dependent Measures.} We use both objective and subjective measures to evaluate agent performance. Objective measures comprised the number of collisions with other vehicles/environment, and the amount of communication (e.g., the number of speech utterances). We also collected a range of subjective measures to ascertain cognitive load, communication properties (helpfulness, redundancy, timeliness, modality selection), and trust after interaction with each agent (See Table \ref{tbl:subjective_qns} in the Appendix). 

\subsection{Results and Analysis}

\begin{figure}
    \centering
    \includegraphics[width=1.0\columnwidth]{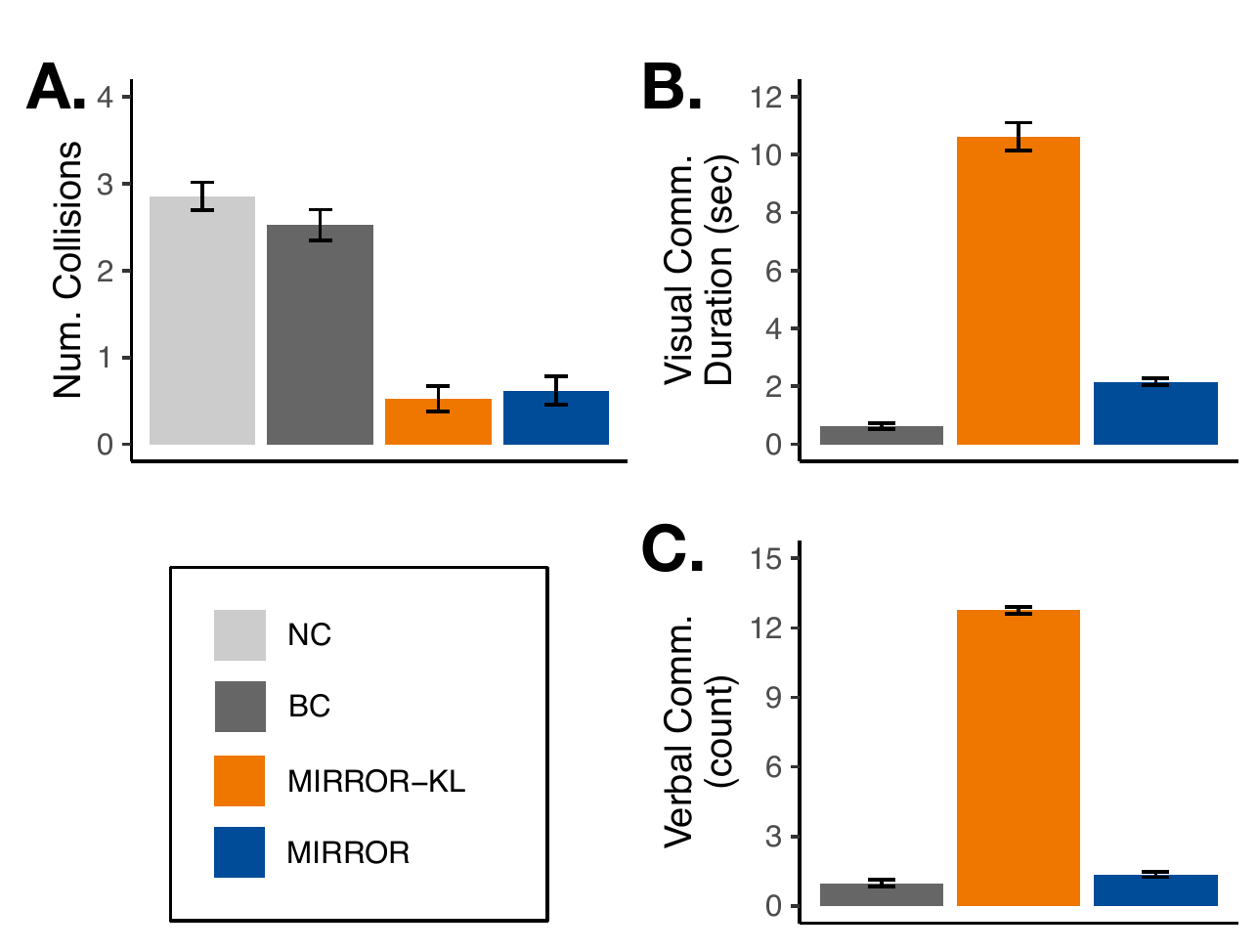}
    \caption{\small Objective Measures. Error bars indicate one standard error. (\textbf{A}) Number of Collisions. (\textbf{B}) Amount of Visual Communication. (\textbf{C}) Amount of Verbal Communication  (\textbf{D}). \mirror{} results in fewer collisions compared to behavioral cloning (\bc) and no communication (\nc) conditions, but with far less communication required compared to belief matching (\mirrorkl).}
    \label{fig:performance_plot}
\end{figure}

In brief, the results support both hypotheses; we give a summary of key findings below and place additional details (e.g., breakdown across NASA-TLX dimensions) in the Appendix. We compared the methods along both objective and subjective dimensions using a repeated measures one-way ANOVA, followed by selected pairwise $t$-tests with adjusted-$\alpha = 0.0167$ using Bonferroni correction. 

Fig. \ref{fig:performance_plot}.A shows that participants experienced significantly fewer collisions when interacting with \mirror{} ($F_{3,60}=47.977$, $p<0.001$; \mirror{} vs \bc{}: $t(9.5)=7.683$, $p<0.001$). Subjectively, participants found \mirror{} provided information that was more helpful and timely compared to \bc{}, and were also more comfortable with the communication modality chosen ($p<0.005$ across the measures and pairwise tests, Fig. \ref{fig:subjective_response} in the appendix). Participants also trusted the \mirror{} agent more than \bc{} ($F_{3,60}=47.730$, $p < 0.001$; \mirror{} vs \bc{}: $t(9.5) = -12.526$, $p < 0.001$). The overall Raw-TLX scores indicated that the participants felt less mentally burdened when they interacted with \mirror{} ($F_{3,60}=10.071$, $p < 0.001$; \mirror{} vs \bc{}: $t(9.5) = 6.658$, $p < 0.001$).  

\begin{figure}
    \centering
    \includegraphics[width=0.80\columnwidth]{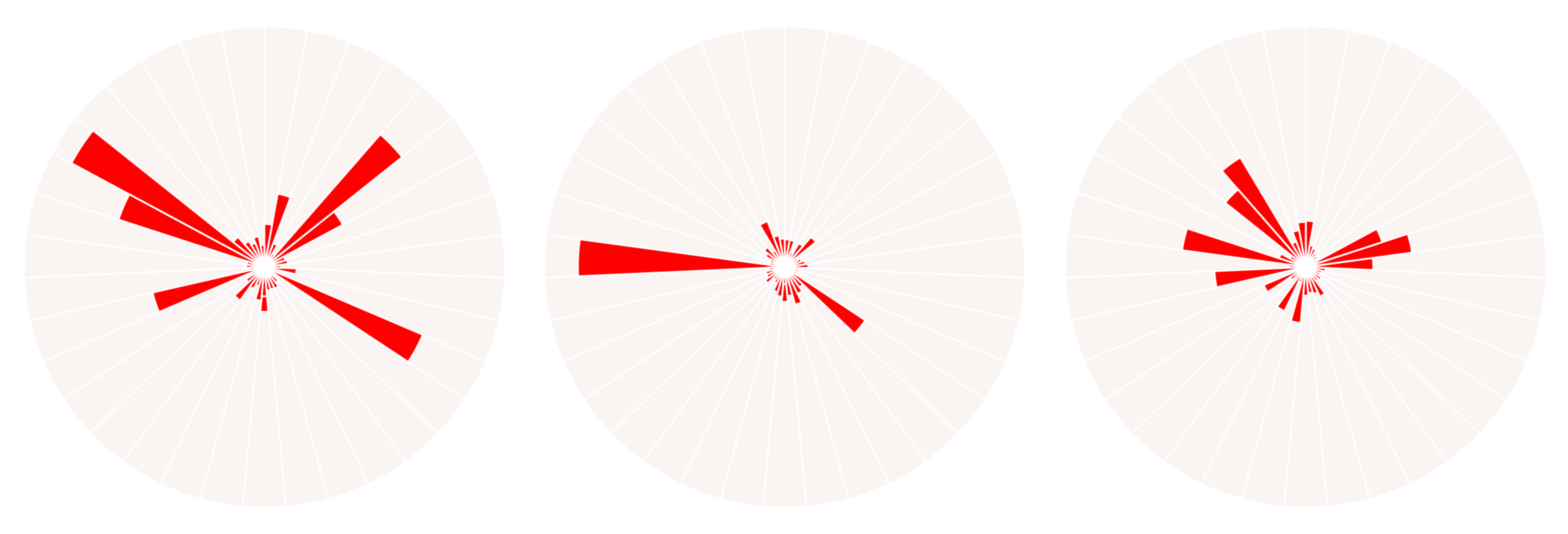}
    \caption{\small Samples of Learned Perceptual Implants (threshold filters) for the CARLA experiment. Top is the front of the vehicle. Length of red bars indicate visibility distance. The implants indicate the human was not able to see far ahead or in the rear but could see cars at the side. Compare against Fig. \ref{fig:carla}.C and \ref{fig:carla}.D.}
    \label{fig:learnedmask}
    \vspace{-1em}
\end{figure}

Taken together, both objective and subjective evidence strongly support hypothesis \textbf{H1}. This finding is corroborated by participant survey responses; they shared that \mirror{} ``\emph{conveys critical information at a good timing}'' in contrast to \bc{}, which they felt ``\emph{is not helping me at all}'' and ``\emph{doesn't inform me about the cars that are approaching from the rear}''. Qualitatively, we found planning with the \bc{} human model to be inaccurate; the \bc{} model would quickly overfit, which led to poor communication. In contrast, the \mirror{} implant models resulted in better communication; Fig. \ref{fig:learnedmask} shows the perceptual implants learned by \mirror{} to well approximate what the human could see, even with  a small amount of training data (six demonstrations).

Next, we turn our attention to \textbf{H2}, i.e., whether 
planning with task rewards and communication costs reduced redundant communication relative to belief matching. Figs. \ref{fig:performance_plot}.B and \ref{fig:performance_plot}.C show how long each agent highlighted cars and the number of times they verbally alerted the driver, respectively. \mirrorkl{} tended to be overly communicative; it provided more than 5 times more visual and verbal communication compared to \mirror{} without significant benefits in terms of task performance. Subjectively, participants rated \mirrorkl{} to provide more redundant information compared to \mirror{} ($t(9.5) = -11.411$, $p<0.001$). In their survey responses, participants wrote that the \mirrorkl{} agent ``\emph{is too talkative}'', ``\emph{told me a lot of useless information}'' and ``\emph{is distracting and annoying}''. In comparison, they found \mirror{} to be ``\emph{straight to the point}.'' Recall that the only difference between the two is the objective function; \mirrorkl{} tries to align belief distributions, regardless of whether the alignment leads to better task accomplishment. We observed that \mirrorkl{} would communicate verbally even when there was no need to, e.g., it would repeatedly tell participants that ``\emph{there is no car in the rear}''.

When asked which agent they were most comfortable with, a majority of the participants (14 out of 21) selected \mirror{}. The remaining participants picked \mirrorkl{}. When asked about which agent they would prefer for long-distance driving (above 1 hour), the number of participants selecting \mirror{} increased to 18. No participant selected \bc{}. Interestingly, some participants preferred \mirrorkl's talkative nature, with one participant stating that they ``\emph{felt annoyed and safe at the same time}''.
A few participants found \mirror{} to be too quiet: ``\emph{I  have some confidence that it works but I'm not entirely sure  because it is quieter}''. These responses suggest individual preferences for information/reassurance and differing trust in the system. How we can incorporate these aspects within \mirror{} would make for interesting future work. 
\section{Conclusion}
\label{sec:conclusion}

In summary, we present \mirror{}, a framework for learning human models using deep self-models for initial structure, along with a planning-based communication approach that couples the human model with learned world model. Experiments show \mirror{} to be effective, outperforming existing behavioral cloning and imitation learning methods. The results show that bootstrapping human model learning with latent-variable models learnt during reinforcement learning leads to generalizable models that are more useful for interaction planning. 

More broadly, we consider \mirror{} to be a step towards more data-efficient human models for human-robot interaction. 
The key idea examined in this work---learning differences from the robot self-model---can be potentially be applied to general human-robot collaboration. Compared to existing work on human models, \mirror{} embodies an alternative paradigm that ``front-loads'' the learning the environmental dynamics and task structure, and thus, offers savings both in terms of sample complexity and computation when learning from human demonstrations. Below, we highlight potential areas for future work: 
\begin{itemize}
    \item In our current setup, the agent and human are not co-present in the environment. We believe \mirror{} can be adapted for these situations by incorporating observations/models of other agents acting in the world. Related work involves crossing the Sim2Real gap, which remains a challenging problem for deep RL.  
    \item Intuitively, \mirror's effectiveness depends on similarity between the human and the robot self-model; drastic differences will render the self-model ineffective as a reference point. We are working on how to characterize this intuition theoretically and potentially estimate in advance how effective \mirror{} might be given varying amounts of data.
    \item A related open question is whether the human model is identifiable within our framework. Similar to IRL, we posit that the differences in the policy and perception cannot be completely ascertained given a dataset. Future work can look into methods that attempt to address this ambiguity, e.g., by  maintaining a distribution over parameters~\cite{ramachandran2007bayesian} or via Bayesian Optimization~\cite{balakrishnan20}. A related issue is how to handle alternative policies and norms that achieve the same objective. 
    \item As our human subject experiments reveal, some human traits  cannot be initially captured within a RL self-model. How we can incorporate elements such as human-robot trust~\cite{kok20trust}, emotions, and social norms remains a key open problem.
\end{itemize}
We believe that building upon \mirror{} forms a compelling pathway towards robots that can interact fluently with humans.

\section*{Acknowledgements}
This research is supported by the National Research Foundation Singapore under its AI Singapore Programme (Award Number: AISG-RP-2019-011).

\balance 
\bibliographystyle{IEEEtran}
\bibliography{references}

\clearpage
\nobalance
\onecolumn
\section{Appendix}

This section contains supplementary material, specifically details on the experiment domains, the simulated human agents, and experimental results relating to the learnt implants. We also provide additional details on the human-subject experiment.

\begin{table*}[bp]
\centering
\caption{Communication Modalities and Costs.}.
\label{tbl:comm_modalities}
{
\small 
\begin{tabular}{l | p{0.15\linewidth} c | p{0.15\linewidth} c}
\hline
\hline 
    \textbf{Domain} & \textbf{Visual} & \textbf{Cost} & \textbf{Verbal} & \textbf{Cost} \\
\hline
    Driving & Relative positions of other cars in front of the agent. & $0.01\cdot(\mathrm{Num.\ cars\ shown})^2$& Symbols representing position and speed of other vehicles. & $0.03\cdot(\mathrm{Num.\ utterances})^2$\\
    \hline 
    Search-\&-Rescue & Raw image pixels revealing a location in the map. Each map is divided into 9 locations. & $0.02\cdot(\mathrm{Num.\ locations\ shown})^2$ & GPT-2 embedding of speech utterances indicating the position of the victim and obstacles, e.g., ``\emph{victim in top-right}''. & $0.1\cdot(\mathrm{Num.\ utterances})^2$\\
     \hline 
    Bomb-Defusal & Raw image pixels of specific terminal. & $0.3\cdot(\mathrm{Num.\ terminals\ shown})^2$ & GPT-2 embedding of speech utterances indicating the bomb type, e.g., ``\emph{Type A Bomb}''. & $0.5\cdot(\mathrm{Num.\ utterances})^2$\\
\hline
\hline
\end{tabular}
}
\end{table*}

\subsection{Domains for Simulated Human Experiment} 
Our experiments made use of three Gridworld domains, each with two communication/observation modalities (see Fig. \ref{fig:exp1_domains} in the main paper and Table \ref{tbl:comm_modalities} below).

\para{Gridworld Driving} In this task, a human drives a car at a constant speed along a road with two lanes. The driver has two possible actions, to stay in his lane or switch lanes. There are four cars around the driver that may speed up or slow down. These cars will not avoid the human's car. The position of the cars are randomly initialized at the beginning of each episode. We examined 8 different scenarios in total with varying car movements. While simple, the state space of this domain is larger than $10^5$. 

In addition to observing their lane, the human driver can receive two types (modalities) of discrete observations (i) the relative position of the other cars and (ii) specific ``speech'' symbols that describe the position of the other cars as well as the speed of the other cars. When driving in clear weather, all cars are visible. However, in dense fog, the driver can only see two units ahead and cannot see cars in the rear. 

\para{Search \& Rescue} In this task, a firefighter has to rescue a victim from a burning house; the firefighter has to find the victim's position and bring them back to the entrance. The victim may appear at one of the three potential positions. There are two corridors to the victim's potential positions, but one of them may be blocked by an obstacle. At the outset, the firefighter is unaware of the positions of victim and the obstacle. We examined 9 different scenarios in total with different situations of corridors and the victim's position. 

The firefighter can receive two kinds of observations in addition to his own position: (i) a raw image of the map and (ii) specific speech utterances that describe the corridors (blocked/unblocked) and  victim's potential position (victim is/isn't there). In the original training (clear) setting, all information are visible. In the transfer setting (dense smoke), the firefighter can only see $1.5$ units around himself and does not observe the speech utterances. 

\para{Bomb Defusal} In this task, a human tele-operates a robot arm to defuse a bomb. To defuse the bomb, the robot needs to press $3$ buttons (a button at each stage). Which button to press at each stage depends on the bomb type and the ``terminals'' alongside the bomb. 

The human can receive two kinds of observations: (i) raw images of specific terminals and (ii) specific speech utterances that indicate the type of the bomb. The robot knowledge of the rules is out-of-date, i.e., its policy is wrong.  As such, the robot cannot defuse the bomb by itself. The human knows the updated rules but is slower than the robot at identifying the correct terminals and is unable to identify the bomb type. 
To help human quickly defuse the bomb, the robot advises the human which button to press and provides explanations (images of specific terminals and the bomb type). The human then chooses a button to press. Note that the human is unable to complete the task on their own since they cannot perceive the bomb type.

\subsection{Simulated Humans Agents.} 
\label{app:real_sim_humans}
In the following, we describe the simulated agents that were created for our experiments. For each domain, we collected data from 10 participants and trained a simulated agent for each participant. The simulated agents are able to perceive state information, up to their perceptual limitations. Qualitatively, we find the behavior of the simulated agents to be very similar to their human counterparts.

\para{Gridworld Driving}
Each participant played 24 rounds in each setting (clear weather and dense fog). 
 We trained a reward function on all the collected human data using Maximum Entropy Inverse Reinforcement Learning~\cite{ziebart2008maximum}. Similar to previous work~\cite{choudhury2019utility}, we use as features (i) the distance to the center of the road, (ii) distances to the other four cars and (iii) driver's action.  Given the learnt reward function, the simulated human agent plans actions at each time-step using CE. Each simulated agent plays 40 rounds (clear weather and dense fog). The last 20 rounds of each simulated agent's data were used as validation set (10 rounds) and testing set (6 rounds). We assumed that the simulated human agent can digest all communicated information and never forgets.

\para{Search \& Rescue}
Each participant played 36 rounds in each setting (clear and smoke). The simulated agent was a planning agent that computes and executes the shortest path between subgoal positions. The subgoals were manually specified, and the transitions probabilities between subgoals were learnt using the collected data. In the smoke setting, if the simulated agent does not know the position of victim, he will search all victim's potential positions until the victim is found. Similar to Gridworld Driving, each simulated agent plays 40 rounds (clear and smoke). The last 20 rounds of each simulated agent's data were used as validation set (10 rounds) and testing set (6 rounds). As before, we assumed that the simulated human agent can digest all communicated information and never forgets.

\para{Bomb Defusal}
Each participant played 36 rounds. For the human participants, we informed them of the type of bomb to ease the number of samples that needed to be collected. 
To determine which button to press, the simulated human has to guess the type of bomb and identify one correct terminal out of the six displayed. 
The type of the bomb is not visible to the simulated human and hence, the chance of guessing the correct type is 0.5. To model the length of time the human takes to find the correct terminal, we use a geometric distribution with success probability $p$ learnt from human data. 
Once the simulated human finds the relevant terminal, it will always press the correct button. Otherwise, it refrains from pressing any button and continue searching for the correct terminal. Same as before, each simulated agent plays 40 rounds. The last 20 rounds of each simulated agent's data were used as validation set (10 rounds) and testing set (6 rounds). We also assumed that the simulated human agent can digest all communicated information and never forgets.

\subsection{\mirror{} Implants} 
\label{sec:exp mirror_implants}

The implants used in our experiments are similar to those described in \ref{subsec:mirror}. 

\para{Perceptual Implants.} In gridworld driving, the perceptual implant is a threshold filter governed by 4 parameters. Each parameter specifies the distance a human can see along a specific direction (front/back and left/right lane) on the road. If a car on a lane is within the specified threshold distance, the model can observe the position of the car. For Search \& Rescue, the perceptual implant is a single parameter threshold filter; we split the image map into 9 portions and if the center of a portion is within this range, it is observable. In bomb defusal, the perceptual implant are 6 parameters --- each models the probability that a human perceives 1 of 6 terminals within 1 second (1 time step). Given these 6 probabilities, we sample a 6 dimensional vector with the value of each dimension to be 0 or 1. If the value of a dimension is 1, we show the corresponding terminal in the image, if not, we mask it out.

\para{Policy Implants.} For all three domains, the policy distributions are modeled as categorical distributions $\mathrm{Cat}(K, \mathbf{p})$. The policy implant was a small neural network that takes in the latent state $z_{t}^\human$ as input and outputs a residual term $\delta(z_{t}^\human)$ that was added to  $\mathbf{p}$ of original policy distribution. The resultant action distribution is then parameterized by   $\hat{\mathbf{p}}=\sigma(\mathbf{p}+\delta(z_t^\human))$.

\begin{figure}
    \centering
    \includegraphics[width=0.5\columnwidth]{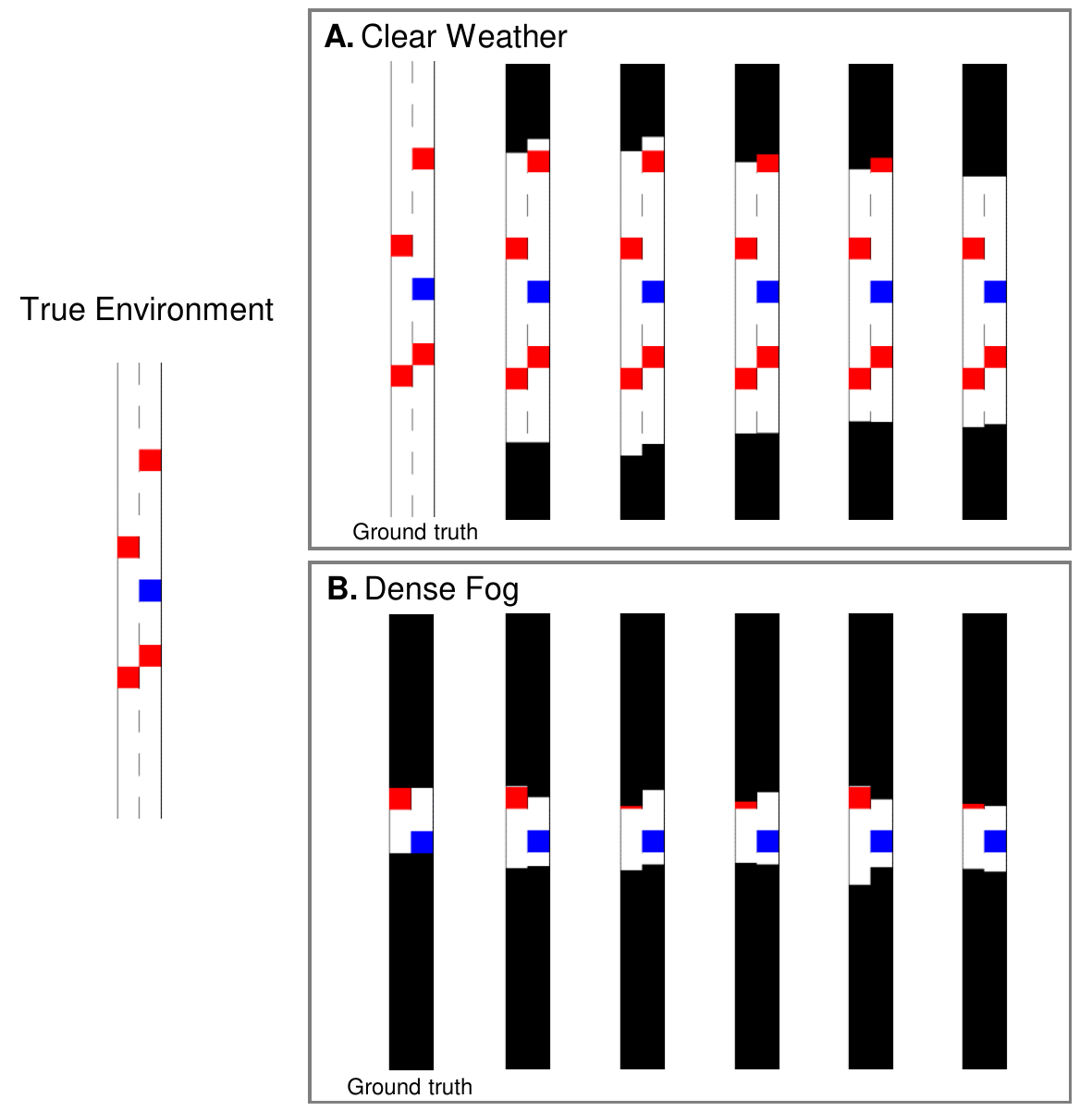}
    \caption{\small Samples of learnt perceptual implants for Gridworld Driving. The blue block represents the ego car and red blocks represent other cars. The black areas represent the region that the human cannot see. Ground truth images represent what human actually sees in the original (clear weather) and transfer (dense fog) settings. (\textbf{A}) The learnt implants indicate that the human was able to see the most of road in clear weather. (\textbf{B}) The implants show the human was only able to see a small nearby area in dense fog. }
    \label{fig:gw_driving:perceptual_implant}
\end{figure}

\begin{figure*}
    \centering
    \includegraphics[width=1.0\textwidth]{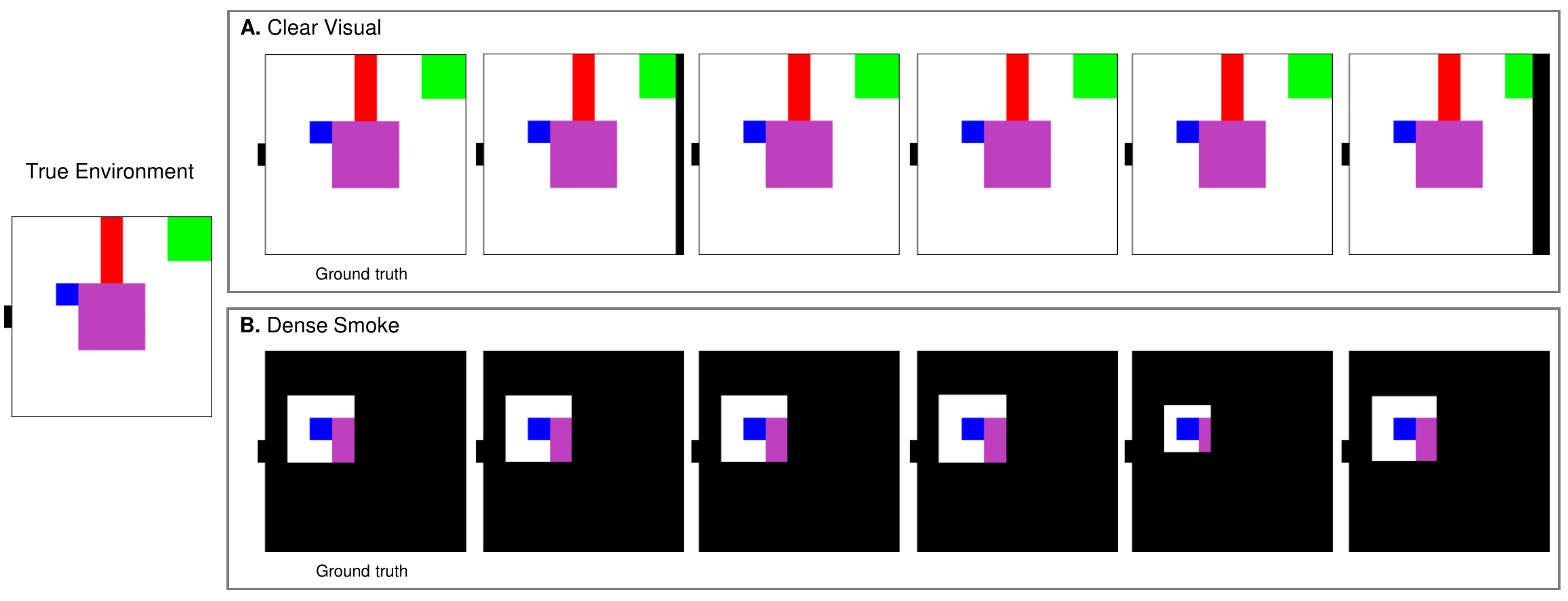}
    \caption{\small Samples of learnt perceptual implants in the Search \& Rescue task. The red and purple blocks represent obstacles and the green box represents the victim. The black areas represent the region that the human can not see. The implants are similar to the ground truth images that represent what human actually sees in the original (clear weather) and transfer (dense fog) settings. (\textbf{A}) For the clear setting, the learnt implants indicate the human was able to see the most of map. (\textbf{B}) In the dense smoke setting, the implants show the human was only able to see a small surrounding region. The implants are similar to the ground truth images that represent what human actually sees in the original (clear weather) and transfer (dense fog) settings.}
    \label{fig:search_rescue:perceptual_implant}
    \vspace{-1em}
\end{figure*}

\para{Learnt perceptual  implants.}
Samples of the learnt perceptual implants by \mirror{} are shown in Fig. \ref{fig:gw_driving:perceptual_implant}, Fig. \ref{fig:search_rescue:perceptual_implant} and Table. \ref{tbl:bomb_defusal:perceptual_implant}). Both in Gridworld Driving and Search \& Rescue, the learnt perceptual implants indicate that the human was able to see in the original setting and was only able to see a small area nearby in transfer setting (e.g., fog). For the Bomb Defusal task, we see the average success rate was approx 0.05. 

\begin{table*}[hbt!]
\centering
\caption{\small $5$ samples of learnt perceptual implants in the bomb defusal game.}

\label{tbl:bomb_defusal:perceptual_implant}
\small  
\begin{tabular}{c c c c c c c}
\hline
\hline 
    Terminals & $1$ & $2$ & $3$ & $4$ & $5$ & $6$ \\
\hline
    Sample $1$  & $0.054$ & $0.051$ & $0.045$ & $0.059$ & $0.185$ & $0.05$ \\
    Sample $2$  & $0.054$ & $0.071$ & $0.044$ & $0.068$ & $0.085$ & $0.068$ \\
    Sample $3$  & $0.045$ & $0.033$ & $0.035$ & $0.045$ & $0.206$ & $0.043$ \\
    Sample $4$  & $0.037$ & $0.034$ & $0.026$ & $0.047$ & $0.012$ & $0.029$ \\
    Sample $5$  & $0.031$ & $0.018$ & $0.026$ & $0.029$ & $0.019$ & $0.037$ \\
\hline
\hline
\end{tabular}
    \vspace{-0.3cm}
\end{table*}

\subsection{Model Architectures and Training}
In all three domains, we approximate the posterior $p(z_{1:T}|x_{1:T}^{1:M}, a_{1:T-1})=\prod_{t=1}^{T}{q_\phi(z_t|z_{t-1}, x_{t}^{1:M}, a_{t-1})}p(z_0)$. Below, we give an overview of our models; source code implementing these models is available at \url{http://blinded-for-review}. 

\para{Gridworld Driving.} Recall that in this domain, state (e.g., position of cars) information is available to the agent. The \mirror{} model consists of the following main components:
\begin{itemize}
	\item The transition distribution $p_\theta(z_t|z_{t-1},a_{t-1})$ and variational distribution $q_\phi(z_t|z_{t-1}, x_{t}^{\mathrm{visual},\mathrm{ verbal}}, a_{t-1})$ are both 48-dimensional Gaussian distribution with diagonal covariance. We set the  transition network to be 3 FC layers deep with hidden size 32. The network parameterizing $q_\phi(z_t|z_{t-1}, x_{t}^{\mathrm{visual}, \mathrm{verbal}}, a_{t-1})$ was a MLP with 3 fully-connected (FC) layers of hidden size 64. 
	\item The observation (decoder) distributions $p_\theta(x_t^\mathrm{visual}|z_t)$, $p_\theta(x_t^\mathrm{verbal}|z_t)$ and reward $p_\theta(r_t|z_t)$ are Gaussian distributions with a diagonal covariance. Each decoder is 3 FC layers deep with hidden size 32. 
	\item The Q function is modeled by a Q-network, which takes $z_t, a_t$ as input and outputs the Q value for the particular $a_t$. The Q-networks consist of 3 FC layers with hidden size 128.
\end{itemize}

The learning rate was set to 0.0003. Before each experience collection phase in Deep Q-Learning, we probabilistically determine whether to use the expert or the model policy via the hyperparameter $\beta$. In our experiments, $\beta$ was initialized to 0.5 with a decay factor of 0.8. Furthermore, whenever the policy is chosen to perform rollout, we perform $\epsilon$-greedy action selection feature to allow for further randomization; $\epsilon$ was initialized to 0.5 with a decay of 0.75 after every rollout phase.

For \bc{}, the policy distribution is a categorical distribution $\mathrm{Cat}(K, \mathbf{p})$. The policy network is modeled as a 48-dimensional GRU followed by one FC layer of size 32; At each time step, the GRU first takes $x_{t}^{\mathrm{visual}}$, $x_{t}^{\mathrm{verbal}}$, $a_t$ and hidden state $h_{t-1}$ from the previous time step as input and outputs a new hidden state $h_t$, which is feed into a fully-connected layer to produce the parameter $\mathbf{p}$ of the categorical policy distribution $\mathrm{Cat}(K, \mathbf{p})$.
For \sqil{}, the Q function is modeled as a combination of a 48-dimensional GRU followed by a 3 FC layers deep with hidden size 32; the GRU first takes $x_{t}^{\mathrm{visual}}$, $x_{t}^{\mathrm{verbal}}$, $a_t$ and hidden state $h_{t-1}$ from the previous time step as input and outputs a new hidden state $h_t$, which is fed into 3 FC layers to produce the Q value.

\para{Search and Rescue}
The distributions and neural networks in the \mirror{} model are similar to the Gridworld driving environment. Accommodating differences in the input (``raw'' image and text) led to differences in:
\begin{itemize}
\item The encoder $q_\phi(z_t|z_{t-1}, x_{t}^{\mathrm{visual}, \mathrm{verbal}}, a_{t-1})$, which uses convolutional layers for the raw image input, and feedforward layers for the symbolic speech observations. Due to computational limits, we pre-trained an autoencoder to reduce the dimensionality of the speech observations (768 dimensional GPT-2 word vector). We set the encoder to have 1 convolutional layer with both kernel size and stride size 3 and outputs 4 channels followed by 3 FC layers with 48 dimensions, while the feedforward portion for the symbolic speech input contains just 1 layer. Each network (Conv for image and FC for speech) produces a 16-dimensional vector, which are concatenated and passed through 3 FC layers (hidden size 128) to derive the state parameters for $z_t$. 
\item The image decoder, which consists of a deconvolution layer with the same kernel size and stride as the convolution layer in the encoder. The symbolic speech decoder has 3 FC layers with hidden size 64. 
\end{itemize}
The Q-network was a MLP with three FC layers (hidden size 200). 
During the training process, $\beta$ was initialized to 1.0 with a decay factor of 0.98 while $\epsilon$ was initialized to 0.5 with a decay factor of 0.9. 

The \bc{} and \sqil{} models are also similar to the Gridworld driving domain, except for the observation input networks. We used the same networks as the \mirror{} model but the concatenated features are fed into a GRU instead of FC layers. 

\para{Bomb Defusal} 
The distribution and model setups were very similar to the above domains; there were only minor differences in the convolutional layers (stride size 2) and FC layers (64 dimensions). The Q network was a larger MLP with 3 FC layers (2048 neurons each). During the training process, $\beta$ was initialized to 1.0 with a decay factor of 0.98 while $\epsilon$ was initialized to 0.5 with a decay factor of 0.9. Likewise, the \bc{} and \sqil{} models were similar to other domains above, but the networks were larger (three layers as before, but with 2048 neurons in each layer). 

\para{CARLA}
The \mirror{} model used higher capacity representations/networks; both $q_\phi(z_t|z_{t-1}, x_{t}^{lidar, verbal}, a_{t-1})$ and $p_\theta(z_t|z_{t-1},a_{t-1})$ were 128-dimensional Gaussian distributions with diagonal covariance. Similarly, the observation distributions ($p_\theta(x_t^{\mathrm{lidar}}|z_t)$, $p_\theta(x_t^{\mathrm{verbal}}|z_t)$), reward $p_\theta(r_t|z_t)$ and policy distributions were Gaussian distributions with diagonal covariance.  The above distributions and the Q-function were modeled using neural networks with 3 fully-connected layers of 128 neurons each. During the training process, $\beta$ was initialized to 1.0 with a decay factor of 0.995 while $\epsilon$ was initialized to 0.5 with a decay factor of 0.9.

For \bc{}, the policy distribution is a Gaussian distribution with a diagonal covariate matrix. The policy network is modeled as a GRU with hidden size 128 followed by a 3 FC layers with hidden size 128; The GRU first takes $x_{t}^{\mathrm{lidar}, \mathrm{verbal}}$, $a_{t-1}$ and the hidden state $h_{t-1}$ from the previous time step as input and outputs a hidden state $h_t$, which is feed into the 3 FC layers to produce the mean and variance of the Gaussian policy distribution. 
 
\subsection{Task Reward}
In the following, we describe the task reward functions used to train our agents and to model the simulated humans.

\para{Gridworld Driving}
If the distance of a car and the ego car is within $2$ units, the agent will receive a $-2$ penalty. If the ego car collides with other cars or goes off the road, the agent will receive a $-5$ penalty. If the ego car changes the lane, the agent will receive a $-1$ penalty.

\para{Search and Rescue}
During each episode, if the human agent finds the victim, they will receive a $+1$ reward. After the agent finds the victim, if they return to the entrance, they will receive another $+15$ reward. Colliding with obstacles incurs a $-10$ penalty. At each time step, the agent will receive a $0.1$ time penalty.

\para{Bomb Defusal}
During each stage, if the human agent press a correct button, they will receive a $+5$ reward. If the agent press a wrong button, they will receive a $-5$ penalty. Each time step incurs a $-1$ cost.

\para{CARLA}
The reward function comprises several components:
\begin{itemize}
    \item Speed reward $r^\mathrm{speed}$, which encourages the agent to drive as fast as possible without exceeding $v_\mathrm{max}$ (set to 40 km/h in the experiments).
    \begin{equation}
        r^\mathrm{speed} = 
        \begin{cases}
        \dfrac{v_\mathrm{current}}{v_\mathrm{max}},& v_\mathrm{current}\leq v_\mathrm{max} \\
        -(v_\mathrm{current} - v_\mathrm{max}),              & \text{otherwise}
        \end{cases}
    \end{equation}
    \item Braking and steering penalties $r^\mathrm{brake}$, $r^\mathrm{steer}$ to promote smooth driving.
    \item Lane change reward $r^\mathrm{change}$, which provides a negative reward of -1 whenever the agent changes a lane, and lane center $r^\mathrm{center}$ reward which penalizes off-center driving using the normalized value of $-\dfrac{\text{distance to the lane center}}{0.5\times\text{ lane width}}$.
    \item Proximity reward $r^\mathrm{proximity}$, which is separated into front and back proximity (penalty of -2 whenever a car is detected within 20 meters by the front and back facing LIDAR beams), and the immediate surrounding proximity (penalty of -4 whenever a car is detected within 1.6 meters by any LIDAR beams)
    \item Road shoulder penalty $r^\mathrm{road}$ of -4 whenever the ego car goes into the road shoulder.
\end{itemize}

{\tiny

\begin{table*}
\begin{center}
\caption{\label{tbl:subjective_qns} Subjective measures in the human experiment.}
\begin{tabular}{@{}ll@{}}
\toprule
& \textbf{Subjective Measures} \\                       \midrule
\begin{tabular}[c]{@{}c@{}}\textbf{After each Method} \\ \textbf{(Including NC)}\end{tabular} &
\begin{tabular}[c]{@{}l@{}}
\textbf{Cognitive Load} \\
\textit{(7-point Likert Scale)} \\ 
- NASA Task Load Index (NASA-TLX)~\cite{hart1988tlx}\\ 
\end{tabular} \\
\midrule
\begin{tabular}[c]{@{}c@{}}\textbf{After each Method} \\ \textbf{(Excluding NC)}\end{tabular} &
\begin{tabular}[c]{@{}l@{}}
\textbf{What, When, How of Human-Robot} \\
\textbf{Communication} \\ 
\textit{(7-point Likert Scale)} \\ 
- The assistive driving agent's communication \\ 
was helpful in accomplishing the task. \\ 
- The assistive driving agent's communication \\
was redundant.\\
- The assistive driving agent's communication \\
was timely.\\
- I feel comfortable with the mode of \\
communication (visual and/or speech) \\
selected by the assistive driving agent at \\
different scenarios.\\
\\
\textbf{Human-Robot Trust} \\
\textit{(7-point Likert Scale)} \\ 
- I trust the assistive driving agent to\\
provide useful communication on this task.\\ 
\end{tabular} \\ 
\midrule
\textbf{After Interaction}  & 
\begin{tabular}[c]{@{}l@{}}
\textbf{Perceived Relative Method Preferences} \\
\textit{(Short Distance vs Long Distance)} \\ 
- Which agent were you the most comfortable with?\\
- Imagine that you have to drive for long distances\\
of at least one hour under the same weather\\
conditions and traffic, which agent would you be\\
the most comfortable with?
\end{tabular} \\ 
\bottomrule
\end{tabular}
\end{center}
\end{table*}

}

\begin{figure*}
    \centering
    \includegraphics[width=0.8\textwidth]{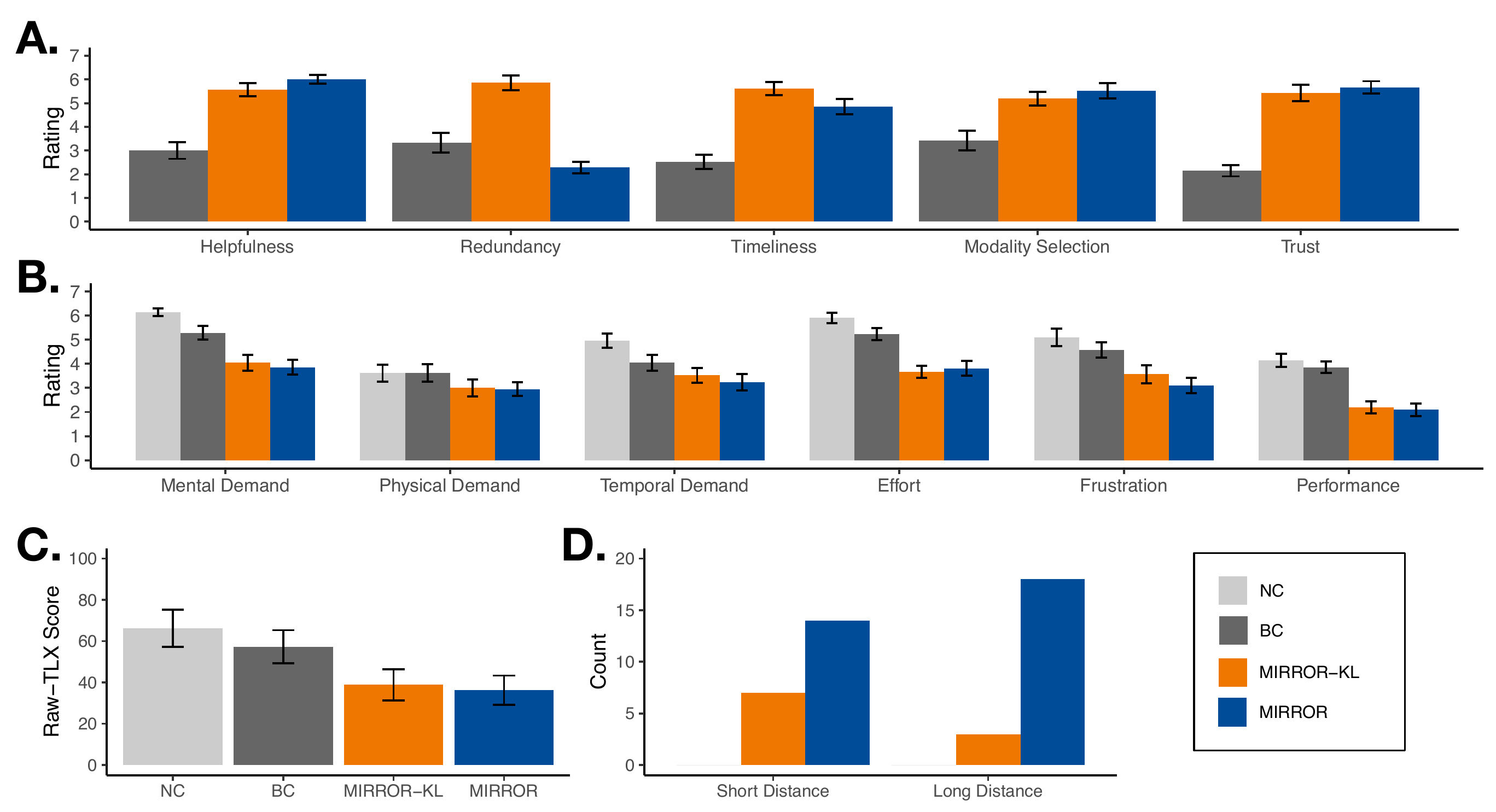}
    \caption{\small Subjective Measures in the CARLA Driving Experiments across conditions. Error bars indicate one standard error. (\textbf{A}) Human-Robot Communication and Trust;  (\textbf{B}) Individual NASA-TLX ratings;  (\textbf{C}) Raw-TLX score; (\textbf{D}) Counts of which agent was preferred for short and long distance driving. Overall, \mirror{} was perceived more positively than \bc{}. Please see the main text for additional details.}
    \label{fig:subjective_response}
\end{figure*}

\subsection{Human-Subject Experiments}
To supplement the main results in the paper, this section provides the specific survey questions (Table \ref{tbl:subjective_qns}) and an analysis of the cognitive workload experienced by the participants.

We focus on the responses collected using  NASA TLX~\cite{hart1988tlx}. Fig. \ref{fig:subjective_response}.B. shows the user ratings for the individual subscales, which supports the notion that \mirror{} lessened cognitive load compared to \bc{}. 
Differences between the communication agents were statistically significant at the $\alpha=0.05$ level across the subscales except for Physical Demand and Temporal Demand; this was likely because the task (driving along a relatively straight highway) was not physically and temporally demanding in nature. Pairwise $t$-tests (adjusted-$\alpha=0.0167$) indicate differences between \mirror{} v.s. \bc{} and \mirrorkl{} v.s. \bc{} to be statistically significant except for Performance (\mirrorkl{} vs BC: $p=0.017$). 

The overall Raw-TLX scores (Fig. \ref{fig:subjective_response}.C.) show that the participants felt less mentally burdened when they interacted with \mirror{} ($F_{3,60}=10.071$, $p < 0.001$; \mirror{} vs \bc{}: $t(9.5) = 6.658$, $p < 0.001$; \mirrorkl{} vs \bc{}: $t(9.5) = 5.580$, $p < 0.001$). \nc{} was always done first (to collect data for training the models), so we cannot completely ignore ordering effects. That said, participants only started the task after sufficient practice. The differences in scores between the \nc{} and \mirror{} conditions are large, which suggests that \mirror{} is effective at reducing cognitive load via assistive communication.

\end{document}